\documentclass[10pt]{article}

\usepackage{standalone}

  \usepackage{amsmath}
   \usepackage{amssymb}
    \usepackage{natbib}
     \usepackage{graphicx}
      \usepackage{amsthm}
      \usepackage{url}
      \usepackage{algorithm2e}
\usepackage{pgfplots}
\usepackage{tikz}
 \usetikzlibrary{arrows,intersections}
 
\usepackage{dsfont}
\usepackage{amsfonts}
\usepackage[scr=boondoxo]{mathalpha}
 
\usepackage[english]{babel}
\usepackage{color}
\usepackage{enumitem}
 \usepackage{thmtools}
\usepackage{thm-restate}
\usepackage{mathtools}
\usepackage{hyperref}
 
\usepackage{cleveref} 

\usepackage{setspace}
\usepackage{algpseudocode}
\RestyleAlgo{ruled}

\usepackage{thm-restate}
 
\usepackage{thmtools}
\usepackage{thm-restate}
\newtheorem{theorem}{Theorem}[section]

 \newtheorem{lemma}{Lemma}[section]
\newtheorem{corollary} {Corollary} [section]
\newtheorem{assumption} {Assumption} [section]
\newtheorem{proposition}{Proposition}[section]

\newtheorem{example}{Example}[section]
 \newtheorem{definition} {Definition} [section] 

\newcommand{\ceil}[1]{\left\lceil #1 \right\rceil}

\DeclareMathOperator{\ir}{int}

\DeclareMathOperator*{\argmin}{argmin}

\DeclareMathOperator{\relint}{relint}

\DeclareMathOperator{\bd}{bd}

\DeclareMathOperator{\dist}{dist}

\DeclareMathOperator{\cA}{\mathscr{A}}
\DeclareMathOperator{\cB}{\mathscr{B}}
\DeclareMathOperator{\cC}{\mathscr{C}}

\DeclareMathOperator{\cM}{\mathscr{M}}

\DeclareMathOperator{\cP}{\mathscr{P}}

\DeclareMathOperator{\cS}{\mathscr{S}}

\DeclareMathOperator{\cX}{\mathscr{X}}
\DeclareMathOperator{\cY}{\mathbb{Y}}
\DeclareMathOperator{\cZ}{\mathscr{Z}}
\DeclareMathOperator{\fC}{\mathfrak{C}}

\DeclareMathOperator{\NN}{\mathbb{N}}

\DeclareMathOperator{\RR}{\mathbb{R}}

\DeclareMathOperator{\sign}{sign}%sign function
\renewcommand{\emptyset}{\varnothing} 

\renewcommand{\leq}{\leqslant}
\renewcommand{\geq}{\geqslant}

\newcommand{\proj}[2]{ {\rm Proj}_{#1}\left(#2 \right)}
\newcommand{\tproj}[2]{{\rm Proj}_{#1}(#2 )}

\newcommand{\norm}[1]{\left\| #1 \right \|}%norm
\newcommand{\tnorm}[1]{\| #1  \|}
\newcommand{\absv}[1]{\left| #1  \right|}%absolute value
\newcommand{\tabsv}[1]{ | #1 |}
\newcommand{\dprod}[1]{\left\langle #1 \right\rangle}%dot-product
\newcommand{\tdprod}[1]{\langle #1 \rangle}
\DeclareMathOperator{\Regf}{{Reg}^f_T}%regret for objective function f
\DeclareMathOperator{\Regc}{{Reg}^c_T}%regret for constraint violations
 \newcommand{\utility}[4]{ \nu^{(#1)}\big(#3_{#2}^{(#1)}, #4_{#2}^{(-#1)}\big)}%untility
  \newcommand{\tutility}[4]{  \nu^{(#1)}(#3_{#2}^{(#1)}, #4_{#2}^{(-#1)})}
   \newcommand{\cf}[2]{\chi_{#1}\left( #2\right)}
     \newcommand{\tcf}[2]{\chi_{#1} ( #2 )}
 \newcommand{\hcf}[2]{\hat{\chi}_{#1}\left( #2\right)}
     \newcommand{\thcf}[2]{\hat{\chi}_{#1} ( #2 )}
 
 \newcommand{\Diameter}[1]{\mathrm{Diameter}\big(#1\big)}
 \newcommand{\tDiameter}[1]{\mathrm{Diameter}(#1)}

\newcommand{\ps}[1]{p_{\rm start}(#1)}
\newcommand{\pe}[1]{p_{\rm end}(#1)}

\newcommand{\R}{\mathbb R}

\newcommand{\oIPM}{online FPM }

\tikzstyle{label}=[rectangle,minimum size=2pt,inner sep=5pt]
\tikzstyle{edge} = [draw,thick,>=latex]
\tikzstyle{pretty} = [circle,thick,draw,fill=black!0, minimum size=15pt, inner sep=0pt]
\tikzstyle{rweight} = [font=\small,xshift=0.4cm]
\tikzstyle{lweight} = [font=\small,xshift=-0.4cm]
\tikzstyle{aweight} = [font=\small,yshift=0.4cm]
\tikzstyle{every edge} = [edge]
\tikzstyle{every state} = [pretty]
\usetikzlibrary{arrows,automata}

 \usepackage{arxiv}

\usepackage{titling}

\thanksmarkseries{arabic}
    
 \newcommand{\shorttitle}{Online Feasible Point Method}
\title{An Online Feasible Point Method for Benign Generalized Nash Equilibrium Problems}
\usepackage{times}
 \author{
 Sarah Sachs \thanks{Bocconi University, 
  \texttt{sarah.c.sachs@gmail.com}}
 \and 
     H\'edi Hadiji 
         \thanks{Univ. Paris-Saclay, CNRS, CentraleSupélec, 
    \texttt{hedi.hadiji@gmail.com}}
    \and
 Tim van Erven  \thanks{University of Amsterdam,
  \texttt{tim@timvanerven.nl}}
  \and
    Mathias Staudigl
    \thanks{Universität Mannheim, 
    \texttt{mathias.staudigl@gmail.com}}
   }

\begin{document}
\maketitle
\begin{abstract}
We consider a repeatedly played generalized Nash equilibrium game. This induces a multi-agent online learning problem with joint constraints. An important challenge in this setting is that the feasible set for each agent depends on the simultaneous moves of the other agents and, therefore, varies over time. As a consequence, the agents face time-varying constraints, which are not adversarial but rather endogenous to the system. Prior work in this setting focused on convergence to a feasible solution in the limit via integrating the constraints in the objective as a penalty function.  However, no existing work can guarantee that the constraints are satisfied for all iterations while simultaneously guaranteeing convergence to a generalized Nash equilibrium. This is a problem of fundamental theoretical interest and practical relevance.
In this work, we introduce a new online feasible point method. Under the assumption that limited communication between the agents is allowed, this method guarantees feasibility. We identify the class of benign generalized Nash equilibrium problems, for which the convergence of our method to the equilibrium is guaranteed. We set this class of benign generalized Nash equilibrium games in context with existing definitions and illustrate our method with examples.

 \end{abstract}

\section{Introduction }\label{sec:intro}
 In this work, we study multi-agent systems with shared constraints. Our goal is
to understand the guarantees with respect to the convergence to an equilibrium and the feasibility of the
repeated interactions between strategic and independent players. If satisfying
the shared constraint is indispensable, it is necessary that the agents
cooperate. To enable this cooperation, we consider a setting where limited communication between agents is possible. 
For each player, this induces an interesting trade-off between
strategic myopic optimization of the losses and cooperation to guarantee
feasibility.  
More precisely, we consider the repeated
interaction of $n$-players on a \emph{generalized Nash-equilibrium problem}
(GNEP) introduced by \cite{63fbe978-3b33-3d32-9230-b215446ef0c4}, see
Definition \ref{GNEP_Def}. The difference to a standard Nash equilibrium
problem is that the players share joint constraints. Hence, the feasible set of
player $i$ may depend on the choices of all other players $x^{(-i)} = [x^{(1)},
\dots, x^{(i-1)}, x^{(i+1)}, \dots x^{(n)}]$.  We define a protocol for the repeated
interaction of $n$-players which allows for some limited communication between
the players. More specifically, before committing to the next iterate
$x_{t+1}^{(i)}$, players are allowed to communicate a \emph{desired set}
$\cS_{t+1}^{(i)}$. These sets indicate a region within which the players want to
play.
 For each iteration $t$, and each player $i$:
 \begin{enumerate}
    \itemsep0em 
    \item Selects a desired set $\cS_{t}^{(i)}$, and sends it to all other players. Receives their sets $\cS_t^{(j)}$ for $j \in [n] \setminus \{ i\}$,
    \item Chooses iterate $x^{(i)}_t$ depending on $\cS_t^{(i)}$ and $\cS_t^{(j)}$'s,
    \item Suffers loss with respect to the loss function. If the shared constraints are violated,  pay $+\infty$. 
 \end{enumerate}
 In this work, we give an algorithm for the updates of the desired sets $\cS_t^{(i)}$ and the iterates $x^{(i)}_t$. 
  Our contributions are as follows:
  \begin{enumerate}
  \item We introduce the \emph{online feasible point method} (online FPM): a new distributed algorithm for computing approximate GNEs.  Under the assumption that each player uses the online FPM in a repeated generalized game, we show in Theorem \ref{feasibilityExploration} that feasibility is guaranteed for each iteration. This online FPM follows a fundamentally different approach from existing methods which are based on Lagrangian splitting schemes or penalty methods. 
  \item We identify a subclass of GNEP, which we name strongly benign GNEP, for which convergence of the iterates to an equilibrium is guaranteed. This is shown in Theorem \ref{thm:convergence}. Furthermore, we illustrate the class of strongly benign GNEP with examples and set it into context with common assumptions such as strong monotonicity for games.
  \item In Section \ref{sec:regret}, Theorem \ref{thm:regret}, we derive regret bounds for each player and set these results in context with existing results on online learning with varying constraints. We note that for online problems with varying constraints, strong guarantees are possible if the constraints are not adversarial but endogenous to the system. 
  \item We illustrate the behavior of the online FPM with various examples of strongly benign GNEP. Furthermore, we demonstrate with experiments that our algorithm convergence to the GNEP without constraint violation for GNEP beyond strongly benign GNEP. 
  \end{enumerate}

 \paragraph{Practical Relevance. }
  A practically relevant example of a GNEP where feasibility is desirable is the
  repeated interaction of agents (sellers and buyers) in the electricity market.
  Most power grids, e.g., the European power grid, impose strong restrictions on
  the fluctuation of the Hertz frequency to avoid risking a blackout. Since a
  nationwide blackout comes at a very high cost for all agents, providing
  feasibility guarantees is indispensable (see \cite{6279592} for more details).
  Further application examples where feasibility guarantees are desirable can be
  found in pollution control or economic models (see \cite{Facchinei_2007} for
  more details).

\paragraph{Related Work.}
 The generalized Nash-equilibrium problem was first introduced by 
\citet{63fbe978-3b33-3d32-9230-b215446ef0c4}. Learning of generalized Nash
Equilibria (GNE) so far relied on a primal-dual approach in which
Lagrangian-based splitting schemes are constructed. This was pioneered by
\citet{Yi2019},  and further studied in a distributed setting \citep{pmlr-v97-liakopoulos19a,8556020,jordan2023first,franci2020distributed}. Penalty
methods are a further toolbox that enforces the feasibility of the limit
solution \citep{facchinei2010penalty, kanzow2019multiplier,Cao21}. 
However, these methods do not necessarily guarantee the feasibility of all iterates in a
distributed setting.   
Our work is related to the flourishing literature on online learning in games \citep{10.5555/2999792.2999954,https://doi.org/10.48550/arxiv.2108.06924,daskalakis2011near-optimal-no}. For zero-sum games without shared constraints, \cite{10.5555/2999792.2999954} provide convergence guarantees for optimistic online methods.  
   In the context of online learning, there is a growing interest in regret guarantees for online learning problems with non-stationary constraints \citep{Neely:2017aa,Liu:2021aa,https://doi.org/10.48550/arxiv.2306.03655}. In contrast to this work, we consider moving constraints that are endogenous to the system rather than adversarially chosen. This makes our setting strictly easier and hence permits us to provide stronger guarantees.

\paragraph{Outline.} 
In Section \ref{sec:GNEP}, we introduce the main objective of our work: the generalized Nash equilibrium problem. We furthermore introduce the subclass of strongly benign and benign generalized Nash-equilibrium problems and set this subclass in context to existing subclasses of generalized games. In Section \ref{onlineIPM}, we introduce our algorithm: an online feasible point method. We illustrate the convergence behavior of this algorithm via experiments on several benign and non-benign generalized games. In Section \ref{Sec:regretBounds}, we analyze the convergence of the online feasible point method. For benign generalized Nash equilibrium problems, we show convergence of the iterates to the equilibrium while preserving feasibility for all iterations. In Section \ref{sec:regret} we relate our findings to regret analysis: we derive regret bounds for the individual players with respect to the loss and the constraint violations. 
 \paragraph{Notation.}
Throughout the paper, we denote indices for iterates by subscripts, e.g., $x_t$ or $\cS_t$, and indices for players by superscripts in brackets, e.g., $x^{(i)}$ or $\cS^{(i)}$.  Following standard notation we denote $[a,b] = \{a,a+1, \dots, b\}$ and $[b] = \{1,\dots , b\}$ for $a,b \in \NN$ with $a\leq b$. Thus, for an $n$-player game $[n] = \{1, \dots, n\} $ denotes the set of indices for all players. 
For sets $\cA,\cB$, we let $\cA + \cB$ denote the Minkowski addition, that is, $\cA + \cB = \{ a+b : a \in \cA, b \in \cB\}$. 
 Following common notational convention, we denote the players' choices of all except the $i^{\rm th}$ player by $x^{(-i)}$ and similarly the product over all sets except the $i^{\rm th}$ by $\cS^{(-i)} = \prod_{j \in [n]\setminus \{ i\}}\cS^{(j)}$. Similarly, we denote  the product space over  all $n$ players by
 $\cS^{[ \times [n]]}_{t} = \prod_{i \in [n]}\cS^{(i)}_t$. Throughout the paper, we let $\norm{\, \cdot\,}$ denote the Euclidean norm. 
 For any compact set $\cZ \subset \RR^n$, we denote the diameter of a set as
  $\Diameter{\cZ } = \max_{x,y \in \cZ}\norm{x-y}$
  and   the distance of any $u \in \RR^n$ to $\cZ$ as 
  $\dist(\cZ,u) = \min_{x \in \cZ}\norm{x-u}$. Further, for compact sets $\cZ$, $\cA \subset \RR^n$, we let $\dist(\cZ,\cA) = \min_{u \in \cA}\dist(\cZ,u)$. 
 Following standard notation convention, we let $\Delta_k \subseteq \RR^k$ denote the $k$-dimensional simplex, that is $\Delta_k = \{ x \in \RR^k : x_i \geq 0, \sum_{i=1}^k x_i = 1\}$.  Further, for any $x \in \RR^d$ and $a>0$, we let $\mathrm{B}_2(x,a) := \{ z \in \RR^d: \norm{x-z}_2 \leq a\}$.

\section{Generalized Nash Equilibrium Problem} \label{sec:GNEP}
In this section, we formally introduce our main objective. Namely, the generalized Nash-equilibrium problem and the subclass of benign generalized Nash-equilibrium problems. 
 \subsection{Setting}
 \label{setting}
  In a repeatedly played generalized Nash-equilibrium problem, each player $i \in [n]$ has control over a variable  $x^{(i)} \in  \RR^{d^{(i)}}$, where $ d^{(i)} \in \NN$ denotes the dimension of the action space of player $i$. We denote $d^{([n])} = \sum_{i \in [n]} d^{(i)}$, and $d^{(-i)} = d^{([n])} - d^{(i)}$.   We let $\fC \in \RR^{d^{([n])}}$ denote the shared constraint set for the players. That is,  we require that $[x^{(1)}, \dots x^{(n)}]^\top \in \fC$. Hence, for player $i$, 
the set of feasible actions, given the choice of actions of the other players, is
 \[
   \cX^{(i)}\big(x^{(-i)}\big) := \big\{ x^{(i)} \in \RR^{d^{(i)}} \mid \big(x^{(i)}, x^{(-i)}\big) \in \fC\big\}.
 \]
 The objective of each player $i \in [n]$ is to minimize their loss $\nu^{(i)}:\RR^{d^{(i)}} \times \RR^{d^{(-i)}}\rightarrow \RR$. 
 With this notation, we introduce the generalized Nash equilibrium problem. The aim of each player $i \in [n]$  in a \emph{Generalized Nash Equilibrium Problem} (GNEP)   is to minimize its loss. That is
 \begin{align*}
 \min_{x^{(i)}}\,  \utility{i}{}{x}{x} \quad \text{ s.t. } \; x^{(i)} \in \cX^{(i)}(x^{(-i)}).
 \end{align*}
Throughout the paper, we assume boundedness and convexity of the constraint set and strong convexity of the loss functions: 
\begin{assumption}
\label{CC}
\begin{enumerate}
\itemsep0em 
\item $\fC$ is non-empty, compact and convex.
\item For all $ i \in [n]$ and $x^{(-i)} \in \RR^{d^{(-i)}}$, the function $\nu^{(i)}(\, \cdot\, , x^{(-i)}) : \RR^{d^{(i)}} \rightarrow \RR $ 
 is differentiable on $\RR^{d^{(i)}}$ and $\mu$-strongly convex. That is, for any $x^{(-i)} \in \RR^{d^{(-i)}}$ it holds for all $x^{(i)}, \tilde x^{(i)} \in \cX^{(i)}(x^{(-i)})$ that 
 \begin{align*}
  \utility{i}{}{x}{x} - \utility{i}{}{\tilde x}{x} \leq \dprod{\nabla_{x^{(i)}}\utility{i}{}{x}{x}, x^{(i)} - \tilde x^{(i)}} - \frac{\mu}{2}\norm{x^{(i)} - \tilde x^{(i)}}^2.
 \end{align*}
 \item For all $i \in [n]$ and $x^{(i)} \in \RR^{d^{(i)}}$, the function $ \nu^{(i)}(x^{(i)}, \, \cdot\,) : \RR^{d^{(-i)}} \rightarrow \RR$ is continuous.
 \item For all $ i \in [n]$, the gradient $\nabla_{x^{(i)}}\utility{i}{}{x}{x}$ is bounded for all $(x^{(i)},x^{(-i)}) \in \fC$, i.e., $\norm{\nabla_{x^{(i)}}\utility{i}{}{x}{x}} \leq G$. We further assume that $G$ is known to the players. 
\end{enumerate}
\end{assumption}
As for standard games, our focus is on computing equilibria. To address approximate generalized equilibria, we adopt the following definition from \cite{Rosen:1965aa}.
  \begin{definition}[Approximate Generalized Nash Equilibrium  (GNE)] 
 \label{GNEP_Def}
  We call $\bar x = [\bar x^{(1)}, \dots, \bar x^{(n)}] \in \RR^{d^{([n])}}$ with $\bar x^{(i)} \in \RR^{d^{(i)}}$   an $\epsilon$--approximate Generalized Nash Equilibrium if for all $i \in [n]$:
 \begin{align*}
 \utility{i}{}{ x}{\bar x} \geq   \tutility{i}{}{ \bar x}{\bar x}-\epsilon\qquad\forall x \in \cX^{(i)}(\bar x^{(-i)}).
 \end{align*}
  If this inequality holds for $\epsilon = 0$, we call $\bar x$ a Generalized Nash Equilibrium.
 \end{definition}
 Due to Theorem~1 in \cite{Rosen:1965aa}, Assumption~\ref{CC} ensures the existence of a GNE. 
 Note that computing the GNE reduces to standard Nash-equilibrium computation if for all $i \in [n]$ the constraints are independent of the opponents choice, that is $ \cX^{(i)}(\bar x^{(-i)}) =  \cX^{(i)}$ for all $x^{(-i)}$. Then $\fC$ reduces to the product space $\prod_{i \in [n]} \cX^{(i)}$. We call any constraint set $\fC$ which can be defined as a product space \emph{uncoupled}, conversely, if this does not hold, we call it \emph{coupled}.  
 
 \paragraph{Benign GNEP.} Let $\phi > 0$. To define the classes of benign and strongly benign GNEP, we define for any $(x^{(i)}, x^{(-i)})$ the set
 \begin{align*}
  \hat{\mathrm{B}}^{(i)}_2(x^{(i)},\phi) :=  \mathrm{B}^{(i)}_2(x^{(i)},\phi)  \cap \cX^{(i)}(x^{(-i)}) \, ,
 \end{align*}
 that is, $  \hat{\mathrm{B}}_2(x^{(i)},\phi) $ gives the $\phi$-neighborhood around $x^{(i)}$, intersected with the set of feasible actions\footnote{Recall from Section \ref{sec:intro} that $\mathrm{B}_2(x,\epsilon)$ defines the $\ell_2$ ball around the point $x$ with radius $\epsilon$.}.
 Based on this set, we define the parameter $ D^{(i)}_{\min}(\phi) $ as
  \begin{align*}
   D^{(i)}_{\min}(\phi):= \max\left\{\alpha \mid \forall x \in \fC, \forall x_+^{(i)} \in   \hat{\mathrm{B}}_2(x^{(i)},\phi) + \left\{-\alpha \frac{\nabla \utility{i}{}{x}{x}}{\norm{\nabla \utility{i}{}{x}{x}}} \right\}:  \left(x^{(i)}_+, x^{(-i)}\right) \in \fC\right\}
  \end{align*}
   and define $D_{\min}(\phi) =    \min_{i \in [n]}D^{(i)}_{\min}(\phi)$. The interpretation of $D_{\min}(\phi)$ is that it gives us the maximal quantity for which the $\phi$-neighborhood around any point in $\fC$ can be shifted in the direction of the (normalized) gradients and still stays inside $\fC$.
   Based on these definitions, we introduce the class of strongly benign and benign GNEP. 
  \begin{definition}[Strongly Benign GNEP]
 \label{def:StronglyBenignGNEP}
  For $\phi, \delta > 0$, we call a GNEP \emph{ $(\phi,\delta)$-strongly benign} if  
 \begin{enumerate}
   \item\label{Benign:uniqueness} The GNEP has a unique GNE $u = (u^{(1)}, \dots, u^{(n)}) \in\relint \fC$ such that for any $i \in [n]$ and $x^{(-i)} \in \cX^{(-i)}$,
   we have $\norm{\nabla \utility{i}{}{u}{x}} = 0$; 
    \item\label{Benign:angularCondition} For all players $i\in [n]$ and all $(x^{(i)}, x^{(-i)}) \in \fC$ with $x\neq u$, it holds that $\nabla \utility{i}{}{x}{x}$ is sufficiently aligned with the vector $x^{(i)} - u^{(i)}$, i.e., there exists $\delta>0$ such that
 \[  \dprod{\nabla \utility{i}{}{x}{x},x^{(i)} -u^{(i)}}   \geq \delta
 \norm{\nabla \utility{i}{}{x}{x}} \norm{x^{(i)} - u^{(i)}}\, \text{;}\]
   \item\label{Benign:NC} $D_{\min}(\phi)   >0$. 
  \end{enumerate}
  \end{definition}
  Note that for $\norm{\nabla \utility{i}{}{x}{x}}\neq 0$ and $x^{(i)} \neq u^{(i)}$, we can rewrite Condition \ref{Benign:angularCondition} as
  \begin{align*}
 \frac{ \dprod{\nabla \utility{i}{}{x}{x},x^{(i)} -u^{(i)}} }{\norm{\nabla \utility{i}{}{x}{x}} \norm{x^{(i)} - u^{(i)}}}  \geq \delta  \, ,
 \end{align*}
that is, we rewrite it as an angular condition on the vectors $\nabla \utility{i}{}{x}{x}$ and $x^{(i)} -u^{(i)}$. Therefore, we will refer to it as the \emph{benign angular condition}.
   \begin{definition}[Benign GNEP]
 \label{def:benignGNEP}
  For $\phi, \delta,\Delta > 0$, 
  we call a GNEP \emph{ $(\phi,\delta, \Delta)$-benign} if it satisfies
  Conditions~\ref{Benign:angularCondition} and~\ref{Benign:NC} from
  Definition~\ref{def:StronglyBenignGNEP}, and, in addition,
  \begin{enumerate}
    \item The GNEP has a unique GNE $u = (u^{(1)}, \dots, u^{(n)})
      \in\relint \fC$ such that for any $i \in [n]$ and $x^{(-i)} \in
      \cX^{(-i)}$, we have $\norm{\nabla \utility{i}{}{x}{x}} \geq \Delta \norm{x^{(i)} - u^{(i)}}$.
  \end{enumerate}
  \end{definition}

\subsection{Relation to common existing assumptions } 
In this section, we set our assumption in context with existing common assumptions.
First, note that under Assumption \ref{CC}, any $ (\phi,\delta)$-strongly benign GNEP is $(\phi,\delta,\mu)$-benign. To see this, note that due to Assumption \ref{CC}, the utilities are $\mu$-strongly convex. Hence, for any strongly benign GNEP
\begin{align*}
&\dprod{\nabla \utility{i}{}{x}{x} - \nabla \utility{i}{}{u}{x}, x^{(i)}-u^{(i)}}\geq \mu\norm{x^{(i)} - u^{(i)}}^2\\
\Rightarrow &\norm{\nabla \utility{i}{}{x}{x} - \nabla \utility{i}{}{u}{x}} \norm{x^{(i)} - u^{(i)}}\geq \mu\norm{x^{(i)} - u^{(i)}}^2.
\end{align*}
Reordering the terms and using that for strongly benign GNEP $\norm{\nabla \utility{i}{}{u}{x}} = 0$ gives the claim. 
To introduce further relations, we recall the definition of quasi-variational inequalities. 
 \begin{definition}[Quasi Variational Inequality]
Let $F:\RR^d \rightarrow \RR^d$ and consider a set valued mapping $C:\RR^d \rightarrow 2^{\RR^{d}}$. For a quasi-variational inequality problem we want to find $x^\star \in  C(x^\star)$ such that
\begin{align*}
\dprod{F(x^\star),x^\star - y} \geq 0 \qquad \forall y \in C(x^\star).
\end{align*}  
We denote this problem by $\mathrm{QVI}(F,C)$
\end{definition}
Suppose $\fC$ is closed and convex and for all $x^{(-i) } \in \RR^{d^{(-i)}}$, $\nu^{(i)}(\, \cdot\, , x^{(-i)})$ is convex, continuous and differentiable, then computing a GNE is equivalent to solving a quasi-variational inequality (QVI) with $F(x) = [\nabla \nu^{(1)}(\, \cdot\, ,x^{(-1)}), \dots, \nabla \nu^{(n)}(\, \cdot\, ,x^{(-n)})]$ and $C (x) = \prod_{i \in [n]} \cX^{(i)}(x^{(-i)}) $. This relation was first discovered by \cite{Bensoussan:1974aa}.   For more details see for example Section 2 in \cite{Pang:2005aa} or Theorem 2 in \cite{Facchinei_2007}. 

Consider a convex set $\fC$ and suppose for all $x \in \fC$, $C(x) = \fC$, then $\mathrm{QVI}(F,C)$ reduces to a variational inequality problem $\mathrm{VI}(F,\fC)$:  
\begin{align*}
\text{find $x^\star \in \fC$, such that } \quad\dprod{F(x^\star),x^\star - y} \geq 0 \qquad \forall y \in \fC.
\end{align*}  
Note that solving a standard Nash-equilibrium problem (NEP) without shared constraints is equivalent to solving a variational inequality problem.  Hence, assuming that $C(x) = \fC$, reduces the GNEP to a standard NEP.  However, there exist sufficient conditions under which the solutions of a QVI are equal to the solutions of a VI.  One such condition is due to a result by  \cite{HARKER199181}. That is, if $A \subset \fC$ such that $\forall x \in A, x \in C(x) \subset A$, then any solution to the $\mathrm{VI}(F,A)$ is a solution to the $\mathrm{QVI}(F,C)$. However, as noted by \cite{Facchinei:2007aa}, in the case of a QVI arising from a GNEP, this condition only allows for uncoupled sets. Hence, under this condition, the problem reduces to an NEP. Note that the strongly benign condition allows for problems that do not satisfy Harker's condition. (See  Section  \ref{sec:examplesBenign} for examples and more details.)
 
A common assumption for variational inequalities is the strong monotonicity assumption: An operator $F: \cX \rightarrow \cX$ is $\mu$-strongly monotone if $\forall x,y \in \cX$
\begin{align*}
\dprod{F(x) - F(y) , x-y} \geq \mu \norm{x-y}^2,
\end{align*}
and strictly monotone if $\tdprod{F(x) - F(y) , x-y}> 0$. 
Strong and strict monotonicity are common assumptions in the context of monotone operator theory or variational inequalities.\footnote{In the context of game theory, strict monotonicity is also referred to as  \emph{diagonal strict concavity}   \citep{Rosen:1965aa}.} In the context of QVI, strong monotonicity is considered in  \cite{Nesterov:2006aa}. In the special case of games, strong monotonicity is for example considered in \cite{Duvocelle_2023,pmlr-v202-yan23f} and referred to as \emph{strongly monotone games}. In the case of $\mu$-strongly convex loss functions (c.f.\,Assumption~\ref{CC}, 2.), the game is $\mu$-strongly monotone. 
We note that under the additional assumption that $\nabla\nu^{(i)}(\, \cdot\,,x^{-i})$ are injective and their inverse is $L^{-1}$-Lipschitz continuous (a.k.a. $L$-bi-Lipschitz), $(\phi,\delta)$-strongly benign GNEP are $\frac{\delta}{L}$-strongly monotone.   Conversely, if the loss functions are $\mu$-strongly convex and $L$-smooth and for GNE $u \in \fC$, $\nabla \utility{i}{}{u}{x} = 0$ then the benign angular condition is satisfied. This is formalized by the following Proposition: 
\begin{restatable}{proposition}{RelationSB}
\begin{enumerate}
\item Consider a GNEP with $\mu$-strongly convex loss functions.  Further, suppose $ \nu^{(i)}(\, \cdot\,,x^{-i})$ are $L$-smooth and there exists a  GNE $u \in  \fC$ such that $\nabla \utility{i}{}{u}{x} = 0$ for all $x \in \fC$. Then the benign angular condition is satisfied with $\delta = \mu L$.
\item Consider a strongly benign GNEP (cf.  Definition~\ref{def:StronglyBenignGNEP}) and assume that the $\nabla\nu^{(i)}(\, \cdot\,,x^{-i})$ are $L$-bi-Lipschitz.   Then the game is $(\delta/ L)$-strongly monotone.
 \end{enumerate}
\end{restatable}
For a proof, see Appendix \ref{appendix:SB}.

 \section{Online Feasible Point Method: Algorithm} 
 \label{onlineIPM}
In this section, we introduce an \oIPM with alternating coordination: the players coordinate by making sure that they update \emph{their desired set} every $n$ turns. The goal of this algorithm is
to guarantee convergence to the GNE while preserving feasibility for all iterates. In Section~\ref{Sec:regretBounds}, we show that this is
guaranteed for an interesting class of games under the assumption that all players use the introduced algorithm.  
Throughout this section, we assume that Assumption~\ref{CC} is satisfied. Recall that the parameter $G$ denotes an upper bound on the gradient norm of the loss functions. Further, we let $D = \tDiameter{\fC, \norm{\, \cdot\,}_2}$.  

\paragraph{A Naive Algorithm: Waiting for Your Turn.}
A first method that fits our framework is to have the players update their 
iterates one player at a time, with every player using an optimization method such as (projected) gradient descent. Formally, player $i$, sends the desired set $\cS_t^{(i)} = \{ x_t^{(i)} \}$ at every turn, and changes the iterate $x_t^{(i)}$ only at timesteps $i + k n$ for $k \geq 0$. 

While this method technically fits our framework, having the players wait for their turn is wasteful as it does not profit from the fact that players may update their plays more often if they declare larger sets $\cS_t^{(i)}$. 
In the rest of this paper, we design and analyze a full algorithm in which the players 
update they play simultaneously while maintaining the joint feasibility thanks to the prior declaration of the desired sets.

 \subsection{Online Feasible Point Method with Alternating Coordination}
 Throughout this section, we assume that all players have access to first-order information, i.e., to gradients $g_t^{(i)}=\nabla_{x^{(i)}}\tutility{i}{t}{x}{x}$. Furthermore, to guarantee feasibility, the players define and communicate the desired sets $\cS_{t+1}^{(i)}$ at the end of each round $t$. In the next round, all players $i \in [n] \setminus \{j\}$ choose their iterate $x^{(i)}_{t+1}$ from $\cS_{t+1}^{(i)}$. The $j^{th}$ player is allowed to update the set $\cS^{(j)}_{t+1}$ and chooses its iterate from this set. 
 To guarantee that all players can make sufficient progress, the player who moves its set varies every iteration. 
 We assume that for each player $i \in [n]$, the online FPM is initialized with $x_1^{(i)}$, $\cS_1^{(i)}$ such that $x_1^{(i)} \in \relint \cS_1^{(i)}$ and $\cS_1^{\times [n]} \subset \fC$. 
 
    \begin{algorithm}[h]
\caption{Online Feasible Point Method with Alternating Coordination }\label{alg:explore}
\textbf{Input: }{$\cS_{1}^{(i)} \subset \RR^{d_i}, x_{ 1}^{(i)} \in \relint \cS_1^{(i)}$}\\ 
\;set $\ps{1} = 1$ and $k = 1$\\
\For{$t = 1, \dots, T$}{
   \If{Termination Criterion \eqref{TC} not satisfied}{
    play $x^{(i)}_{t}$, suffer loss and receive gradient $g_t^{(i)}$\\
     update $ x^{(i)}_{t+1}$ according to \eqref{update}\\
      update $\cS_{t+1}^{(i)}  $  according to \eqref{Vupdate}\\    
    send $\cS_{t+1}^{(i)}$ to all players $j \in [n]\setminus \{i\}$ and receive $\cS_{t+1}^{(j)}$ from all $j \in [n] \setminus\{i\}$
}
\If{Termination Criterion \eqref{TC} satisfied}{
     set $x_{t+1}^{(i)} = x_t^{(i)}$\\
     shrink $\cS_t^{(i)}$ such that $x^{(i)}_t \in \relint \cS_{t+1}^{(i)}$ and $\Diameter{\cS_{t+1}^{(i)}} = \frac{1}{2} \Diameter{\cS_{t}^{(i)}}$\\
      set $\pe{k} = t$ and $\ps{k+1} = t+1$ and increment $k$ by one
     }
}

\end{algorithm}
The online FPM proceeds in phases. The $k^{\mathrm{th}}$-phase starts at $\ps{k}$ and ends at $\pe{k}$.  We note that $\ps{k}\leq\pe{k}$ and denote $\cP_k = \{\ps{k},\dots,\pe{k}\}$. 
It remains to specify the termination criterion \eqref{TC} and the updates of the iterates $x_t^{(i)}$ and $\cS_t^{(i)}$, \eqref{update} and \eqref{Vupdate} respectively.   
 The termination criterion is triggered if none of the players can make sufficient progress or if it was not triggered for $2^k$ iterations. That is, 
\begin{align*}
\label{TC}
 \max_{i \in [n]} \dist\left(\cS^{[\times [n]]}_{t-i },\cS^{[\times [n]]}_{t-i+1} \right)\leq  \frac{D}{\sqrt{T}} \qquad \text{or} \qquad t-\ps{k} \geq  2^k \tag{\textbf{TC}}.
\end{align*}
 To define the updates of $x_t^{(i)}$ and $\cS_t^{(i)}$, we let $\eta >0$ denote a fixed step size to be defined later. The updates are done with respect to the minimum between $\eta$ and the parameters $\iota_t\geq0$ and $\bar \eta_t/2>0$ to guarantee feasibility. To define $\iota_t$, set $\cZ^{(i)}_t =\{z : z \in \argmin_{z \in \cS_t^{(i)}} \tdprod{g_t^{(i)},z}\}$. Define \[\iota^{(i)}_t = \min \dist(\cZ^{(i)}_t \times \cS_t^{(-i)}, \bd(\fC)).\] That is, $\iota_t^{(i)}$ is the minimal distance with respect to the gradient direction $g_t^{(i)}$ of the set $\cS^{\times [n]}_t$ to the boundary of $\fC$. 
  For the updates of the set $\cS_t^{(i)}$, we set
 \begin{align}
 \cS_{t+1}^{(i)} = \begin{cases}
 \cS_t^{(i)}&\text{ if } t \bmod n \neq i-1\\
 \cS_{t}^{(i)} + \{ v_t^{(i)}\}&\text{ if } t \bmod n = i-1
 \end{cases},\label{Vupdate} \tag{\textbf{Vupdate}}
 \end{align}
 where $v_{t}^{(i)} = \min(\eta, \iota_t^{(i)}) g_t^{(i)}$. 
  Further, we set $\bar \eta_t^{(i) } = \max(\alpha \geq 0: (x^{(i)}_t - \alpha g^{(i)}_t) \in \cS^{(i)}_t)$, that is, $\bar \eta_t^{(i)}$ is the maximal step into the direction of $g^{(i)}_t$ while staying within $\cS_t^{(i)}$.  
Define the update 
\begin{align}
 x_{t+1}^{(i)} = \begin{cases}
 x_t^{(i)} -  \min(  \eta, \bar\eta_t^{(i) }/2 ) \,g_t^{(i)} &\text{ if } t \bmod n \neq i-1\\
 x_t^{(i)} -  \min(   \eta,\iota_{t}^{(i)})  \,g_t^{(i)} &\text{ if } t \bmod n = i-1
 \end{cases}\label{update} \tag{\textbf{Update}}
 \end{align}

 \section{Online Feasible Point Method: Feasibility and Convergence }
\label{Sec:regretBounds} 
In this section, we show that the \oIPM   is guaranteed to preserve feasibility. Furthermore, we show that for strongly being GNEP with $L$-smooth and $\mu$-strongly convex loss functions (cf.\, Assumption \ref{CC}), convergence to the GNE is guaranteed. 
\begin{restatable}{theorem}{feasibilityExplor}\emph{[Feasibility]}
  \label{feasibilityExploration}
  Suppose all players are following Algorithm \ref{alg:explore}. Then for all iterations $t \in [T]$, we have $\cS^{[\times [n]]}_t \subseteq \fC$; this implies in particular that $[x_t^{(1)},\dots, x_t^{(n)}] \in\fC $.
 \end{restatable}

 This result is a straightforward consequence of the definition of the algorithms, for a formal proof see Appendix \ref{proofs:algo}. Furthermore, for strongly benign and benign GNEP, the online FPM converges to the unique GNE. 

  \begin{restatable}{theorem}{oFPMconvergence}[Convergence]
  \label{thm:convergence}
   Suppose Assumption \ref{CC}  is satisfied and assume all players are following Algorithm \ref{alg:explore} with step size $\eta = \min(\frac{D}{G\sqrt{T}}, \frac{\delta}{L})$. 
 Assume we have a $(\phi,\delta)$-strongly benign GNEP.  Set \[t_0  =  \max\left(   \frac{4D}{\phi} +1,\;  \left(\frac{D}{2D_{\min}}\right)^2, \; \left(\frac{DL}{2G\delta} \right)^2\right).\]  
  Then for all players $i \in [n]$  and any $t  \in[ t_0, T]$ 
    \begin{align*}
  \norm{x_t^{(i)} - u^{(i)}} \leq  \Xi \norm{x_1^{(i)} - u^{(i)}}  \left(1-\frac{ \mu \delta D}{4 G\sqrt{T}}\right)^{  \frac{ t+1}{n}}
     +  \frac{2D}{\delta \sqrt{T}} ,
  \end{align*}
  where
   $  \Xi =   \left(1-\frac{ \mu \delta D}{4 G\sqrt{T}}\right)^{  -\frac{ t_0}{n}}$.
      
 \end{restatable}
 For a complete proof and a similar result for benign GNEP, see Appendix \ref{proofs:regret}. Our proof relies on a special case of an inexact gradient descent analysis. That is, a gradient descent scheme of the form $x_{t+1} = x_t - \eta_t g_t$ where $g_t$ does not necessarily correspond to the gradient, but is a vector that is sufficiently aligned with the real gradient. This alignment is guaranteed for benign GNEP via the benign angular condition.  
 Hence, the benign angular condition is essential for this analysis since it ensures that the gradients of the loss function are relatively well aligned with the gradients of the function $f(x) = 2\delta^{-1}\norm{x-u}$.

  \subsection{Comparison to Alternating Gradient Descent}
  \label{compAGD}
  As mentioned in Section \ref{onlineIPM},
 an alternative approach to the online FPM (Algorithm \ref{alg:explore}) would be an alternating gradient descent method. That is, for each step $t \in [T]$, each player $i\in[n]$ sets \begin{align}\label{AGD}
 x_{t+1}^{(i)} = \begin{cases}x_t^{(i)} - \eta_t g_t^{(i)} &\text{ if } t \bmod n = i-1\\
 x_t^{(i)}&\text{otherwise}.
 \end{cases} \tag{Alt-GD}
 \end{align}
This method has the clear advantage of being easier to implement and requiring no extra storage for the sets $\cS_t^{(i)}$.  Note that the feasibility of the iterates can be guaranteed if the parameter $D_{\min}(0)$ is available for stepsize tuning or all $x_t^{(j)}$ of all players $j \in [n]\setminus\{i\}$ are known to player $i \in [n]$. Indeed, in the latter case, feasibility can be guaranteed by a projection onto $\cX^{(i)}(x_t^{(-i)})$ and in the former case, feasibility with $\eta = D_{\min}(0)/(\sqrt{T}G)$ follows from the definition. Thus, assuming that either of these parameters is available can also guarantee feasibility for this algorithm. 
However, note that alternation is essential for this argument; therefore, a player can only make progress every $n^{th}$-iteration. 
Conversely, note that with the online FPM, Algorithm \ref{alg:explore}, the players update their iterates every iteration while guaranteeing feasibility. While this does not always help, in some cases of practical interest it can lead to a significant speedup as the following example shows:
\begin{example}
Consider a GNEP with $\fC = \prod_{i \in [n]} \Delta_{d}$ and $\utility{i}{}{x}{x}$ such that there exists a unique GNE $(\bar x^{(1)}, \dots, \bar x^{(n)})$ in the strict interior of $\fC$. 
Comparing the online FPM (Algorithm \ref{alg:explore}) with the alternating gradient descent method \eqref{AGD}, we note that the players can update their iterates in the online FPM every iteration, while for \eqref{AGD}, they only make progress every $n^{th}$ iteration. Furthermore, for this specific problem, given that $\bar x \in \ir\left( \cS_t^{(\times [n])} \right)$ every player makes progress towards the GNE with a step size of at least $\min(\Diameter{\cS_t^{(i)}}/(2G)  , \eta)$. Given that $\Diameter{\cS_t^{(i)}}$ is a constant, each player makes constant progress towards the GNE.  Specifically in the case when the number of players is large, e.g., proportional to $\sqrt{T}$, this speed-up can be significant and of practical interest.
\end{example}

Another advantage of the online FPM is its robustness beyond the theoretical guarantees. As our experiments show (cf. Appendix \ref{ExAndEx}), the online FPM can work well beyond the theoretical guarantees. In particular, it always guarantees feasibility and convergence to GNE for instances beyond the class of benign GNEP. Capturing the described advantages of the online FPM in a rigorous analysis constitutes an interesting future research direction (see Section \ref{future} for more details).
  
 \section{Relation to Regret Minimization}\label{sec:regret}
In this section, we highlight the relation of our algorithm to no-regret algorithms in repeated games.    
 From the perspective of a single player, interacting in a game with shared
constraints can be viewed as a convex online learning problem with time-varying constraints. Note that in contrast to standard online learning problems, where the constraint is fixed,  online learning with time-varying constraint sets is a fundamentally more challenging setting. In recent years this setting received more interest in the online learning community (see e.g.\,\cite{Neely:2017aa,Liu:2021aa,https://doi.org/10.48550/arxiv.2306.03655}). 

 Consider the following online learning protocol with time-varying constraints: For each round $t \in
[T]$, the learner chooses an iterate $x_t \in \cX$, receives convex functions
$f_t: \cX \rightarrow \RR$ and a convex constraint set $\cC_t = \{x \in\cX\mid
h_t(x) \leq 0 \} \subset \cX$ with $h_t:\cX \rightarrow \RR$ and suffers
instantaneous regret with respect to $f_t(x_t)$ plus a  cost for
violating the constraint $\cC_t$.
A common evaluation criterion for the performance of an algorithm for an online
learning problem is the \emph{regret}. For
any $u \in \cX$, define
\begin{align}
\label{def:regretF}
	\Regf(u) := \sum_{t=1}^T  f_t(x_t) -\sum_{t=1}^T f_t(u) .
\end{align}
An online algorithm is said to be  a \emph{no-regret algorithm} if for any $u \in \cX$, $\Regf(u)$ grows sub-linearly in $T$.  Since the algorithm faces time-varying constraints, we also want to measure the performance with respect to the constraint violations. 
  There exist several definitions of regret
for constraint violation in the literature. For example, \cite{JMLR:v13:mahdavi12a} use the regret definition $  
\sum_{t=1}^T(h_t(x_t) - h_t(u))$   and  \cite{10.5555/3327345.3327512} define the stronger version $  \sum_{t=1}^T(
h_t(x_t))_+ - (h_t(u))_+)$   \footnote{Note that
if $ \emptyset \neq\cap_{t \in [T]} \cC_t \ni u$, this reduces to $\sum_{t=1}^T
(h_t(x_t))_+$. } to measure the constraint violations.
There also exist last step requirement like $ h_T(x_T) \leq O(1/\sqrt{T})$ as used in 
\cite{https://doi.org/10.48550/arxiv.2306.03655}.  

 To capture our results in terms of regret, we introduce a definition of regret which is based on the indicator function. That is, let $\chi_{\cC_t}:\cX \rightarrow
\RR \cup \{ + \infty\}$  denote the characteristic function with $\tcf{\cC_t}{x} = 0$
if $x \in \cC_t$ and $+\infty $ otherwise. Define the  regret with respect to
constraint violations
\begin{align}
\label{def:regretC}
\Regc = \sum_{t=1}^T \cf{\cC_t}{x_t} .
\end{align}
Note that $\Regc$ is equal to the \emph{dynamic regret} $ \sum_{t=1}^T \cf{\cC_t}{x_t}  - \cf{\cC_t}{u_t}$ where $u_t \in   \cC_t$. Furthermore, $\Regc \in \{0,+\infty\}$. Hence, requiring an algorithm to be a no-regret algorithm with respect to  $\Regc$ is equivalent to requiring $\Regc =0$. 
We note that for any sequence $(x_t)_{t\in[T]} \in \cX^T$ and $u \in \cX$
\begin{align*}
\sum_{t=1}^T \cf{\cC_t}{x_t}\geq \sum_{t=1}^T(
h_t(x_t))_+ - (h_t(u))_+).
\end{align*}
Hence, if an algorithm is no-regret with respect to $\Regc$, then $ \sum_{t=1}^T(
h_t(x_t))_+ - (h_t(u))_+)$  is non-positive. Note that the converse is not necessarily true. 
Furthermore, if $\Regc = 0$, this immediately implies that $h_t(x_t) \leq 0$ for all $t \in [T]$. Hence, we obtain $h_T(x_T) \leq c\frac{1}{\sqrt{T}}$, which is a requirement considered in \cite{https://doi.org/10.48550/arxiv.2306.03655}.
 
 From Theorem \ref{thm:convergence} we derive sublinear regret bounds. We adapt the notation to be consistent with the previous notation: the time-varying convex functions are $ \nu^{(i)}( \, \cdot\,,x_t^{(-i)})$ and the time-varying constraint sets are $\cX^{(i)}(x^{(-i)}_t)$. Note that both are time-varying due to the choice of $x_t^{(-i)}$. 
\begin{restatable}{theorem}{regretBoundSB}
\label{thm:regret}
Suppose Assumption \ref{CC} is satisfied and assume we have a $(\phi,\delta)$-strongly benign GNEP. 
Let $\Xi$ and $t_0$ be defined as in Theorem  \ref{thm:convergence}. If all players are following Algorithm \ref{alg:explore}, then for all players $i\in[n]$
\begin{align*}
\sum_{t=1}^T \cf{\cX^{(i)}(x^{(-i)}_t)}{x^{(i)}_t}=0,
\end{align*}
and for all $T \in \NN$
\begin{align*}
\sum_{t=1}^T   \nu^{(i)}\left(x_t^{(i)},x_t^{(-i)}\right)  -\sum_{t=1}^T    \nu^{(i)}\left( u^{(i)},x_t^{(-i)}\right) \leq  DG\left( \sqrt{T}\left(2 \Xi n\frac{G}{\mu D} + \frac{2}{\delta} \right) + t_0\right) . 
\end{align*}
 \end{restatable}
This result is a direct consequence of the convexity of the losses and Theorem~\ref{thm:convergence}. For a proof, see Appendix \ref{appendix:regretBounds}.

From a single-player perspective, the online learning problem with time-varying constraints induced by a GNEP is fundamentally easier than an online learning problem with adversarially time-varying constraints. The following three aspects are essential: (1) the constraints are not adversarial but endogenous to the system, (2) before committing to the next iterate, the player has access to side information $\cS_t^{(i)} \subseteq \cX^{(i)}(x_t^{(-i)})$, and (3) the distance between $u^{(i)}$ and $\cS_t^{(i)}$ is decreasing. 
We note that if the time-varying constraints are chosen adversarially, it is not possible for any online
algorithm to guarantee that $\Regc$ is bounded\footnote{ For example,
consider $\cX = [-1,1]$. For any sequence of $\{x_t\}_{t \in [T]}$ there exists
a sequence of adversarial time-varying constraints $\{\cC_t\}_{t\in [T]},
\cC_t \subset \cX$ such that $\emptyset \neq \cap_{t=1}^T\cC_s$ and at least
one $x_t \notin \cC_t$. This follows due to the density of the reals. Hence,
for any $u \in \cap_{t=1}^T\cC_s$, $\Regc(u) = \infty$.}.
 Based on these insights, we deduce an online learning protocol where the learner receives a set $\cS_t \subseteq \cC_t$ as side information before he has to choose his next iterate.
The protocol is as follows: In each round $t \in [T]$, the learner receives a
set $\cS_t \subseteq \cC_t$ and then chooses the iterate $x_t$. Next, the
player learns $f_t$ and $\cC_t$ and suffers a loss with respect to $f_t$. If
the player violates the constraint $\cC_t$, he suffers a loss of $\infty$.

By choosing its iterates $x_t \in \cS_t$, the learner can guarantee that the
constraints $\cC_t$ are never violated. However, it is not necessarily possible
to simultaneously guarantee sub-linear regret for $\Regf(u)$.  For illustration, we consider the following simple example: Suppose the relative interior of $\cap_{t=1}^T
\cC_t$ is non-empty and take any $x,y \in \cap_{t=1}^T\cC_t$ with
$\dist(x,y)>0$ a constant. Suppose the learner is provided $\cS_t = \{x\}$ for $t
\leq T/2$ and $\cS_t = \{y\}$ otherwise.  Let  $u \in \cap_{t=1}^T\cC_t $ and define $f_t(x) = a_t \norm{x-u}^2$ for any $a_t>0$. Then
$\Regf(u)$ is linear in $T$ whenever the player chooses $x_t \in \cS_t$ to avoid constraint violations. 
Based on these insights, we derive the following result: Suppose the learner obtains sets $\cS_t \neq \emptyset$ with $\cS_t\subset \cC_t$. Assume that the distance $\dist(\cS_t,u) \leq \frac{c}{\sqrt{t}}$ and the distance between the sets $\cS_t$ and $\cS_{t+1}$, i.e., $\max_{a \in \cS_{t}}\dist(\cS_{t+1},a)$, is uniformly bounded for all $t \in [T]$.  Assume the functions $f_t:\RR^d \rightarrow \RR$ are convex and differentiable. Then the regret for projected online gradient descent, i.e., $x_{t+1} = \tproj{\cS_{t+1}}{x_t - \eta_t\nabla f_t(x_t)}$, is sublinear while the constraints are never violated. That is, 
   \[
    \Regf(u)
      \leq C\frac{3D + 10c}{2}G\sqrt{T} \quad \text{and}\quad \sum_{t=1}^T \cf{ \cC_t}{x_t}=0,
  \]
  where  $D \geq \Diameter{\cS_t}$, $G \geq \tnorm{\nabla f_t(x_t)}$ and $C>0$ denotes a constant independent of $T$. See Lemma~\ref{regretSinglePlayer} in Appendix~\ref{COCO} for more details.

  \section{Discussion and Future Work}
 \label{future}
 \paragraph{Online FPM with Simultaneous Coordination.} The alternating coordination in Algorithm \ref{alg:explore} requires that the players are enumerated and know their identity $i$. Otherwise, it is impossible to determine whether $t\bmod n = i-1$.  
 Furthermore, the progress of a player during iterations where $t\bmod n \neq i-1$ can be negligible. Due to these limitations, we note that an online FPM with \emph{simultaneous} shifts of the sets $\cS_t^{(i)}$ for all players $i\in[n]$ might be of practical and theoretical interest.   
 \paragraph{Extension of the Theoretical Analysis beyond Benign GNEP.}
 As our experiments illustrate, the online FPM converges to a GNE for GNEP beyond the restrictive class of benign GNEP. We leave it to future work to identify further subclasses of GNEP for which convergence can be guaranteed. Identifying subclasses of GNEP with practical relevance and extending the analysis is an interesting future research direction. 
 \paragraph{Better Convergence Guarantees.} Note that our convergence analysis does not reflect the progress made by the players in every step $t: t \bmod n = i-1$. As we illustrated in Section \ref{compAGD} via an ad-hoc argument, this progress can lead to better convergence in practice. However, this is not reflected in our analysis. Improving the analysis such that it can capture these steps constitutes an interesting future research direction. 
\paragraph{Simultaneous cooperation and non-cooperation games.}
 Requiring that the players never violate a constraint in a repeated GNEP compels cooperation between the players. At the same time, all players have a selfish interest in minimizing their losses. This can be interpreted as a simultaneous cooperation and competitive game: The competitive aspect of the game is captured by the requirement to optimize the payoffs given by the loss function $ \utility{i}{}{ x}{x}$ in every round. To account for the cooperation aspect, we define a cooperation game based on the constraint violations. A cooperation game with $[n]$ players is defined via a characteristic function that gives the collective payoff for a coalition between $\cM$ players with $\cM \subset [n]$.  Hence, we define a finite variant of the indicator function $\thcf{\cC }{\, \cdot\,} : \cX \rightarrow \RR$ with $ \thcf{\cC }{x} = 0$ if $x\in \cC$ and otherwise $\thcf{\cC }{x} = C$ where $C>0$ denotes a (large) constant. We let $\psi:2^{[n]} \rightarrow \RR\ $ with $ \psi(\emptyset)=0$ and 
 \begin{align*}
 \psi(\cM) = C -  \frac{1}{\absv{\cM}} \sum_{j\in\cM} \hcf{\cX^{(j)}(x^{(-j)})}{x^{(j)}}. 
 \end{align*}
The players can guarantee that this characteristic function $\psi$ is $C$ by forming the \emph{grand coalition}, that is the coalition consisting of all players $[n]$. Choosing $C$ sufficiently large guarantees that the players have an incentive to form a grand coalition. Note, however, that $\psi$ defines a very simplistic cooperation game.  Exploring this connection further might be an interesting future research direction.  

 \paragraph{Acknowledgements.}	
 We thank Crist\'obal Guzm\'an for the discussion during the preliminary phase of this project. 
 Van Erven was supported by the Netherlands Organization for Scientific Research (NWO) under grant number VI.Vidi.192.095. This research was performed while Sachs was at the University of Amsterdam. During that time, she was supported by the same grant number VI.Vidi.192.095. Staudigl acknowledges support from the COST Action CA16228 ”European Network for Game Theory”.
  
 \bibliographystyle{apalike}
 \bibliography{Literatur}{}

\appendix
\section{Examples and Experiments}
\label{ExAndEx}
\subsection{Examples and Experiments for Benign GNEP}
  \subsubsection{Examples of Benign and Strongly Benign GNEP}\label{sec:examplesBenign}
  The conditions for benign and strongly benign GNEP are strong and exclude many instances of GNEP. However, it is a nontrivial subclass of GNEP as we illustrate in the following examples.    The first example serves as an illustration of the definitions and is relatively trivial. The reason is that the example is one-dimensional. This enables us to illustrate it via plots. However, conditions such as the benign angular conditions constrain us to trivial uncoupled loss functions in one dimension. 
 \begin{example}[$(\phi,\delta)$-strongly benign GNEP]\label{Example:SB}
 Consider a two-player game where $n_1 = n_2 = 1$. For simplicity, we denote the players action by $x \in \RR$ and $y \in \RR$ instead of $x^{(1)}$ and $x^{(2)}$. Furthermore, we denote the sets $\cS_t^{(x)}, \cS_t^{(y)}$ instead of $\cS_t^{(1)}, \cS_t^{(2)}$. 
 Define the shared constraint set
 \begin{align*}
 \fC = \{ (x,y) \in \RR^2 : 8 \geq x \geq -1,\; 8\geq y\geq-1, \;x+y \leq 10  \}.
 \end{align*}
 The players want to minimize their losses with respect to $\fC$. That is
 \begin{align*}
 \min_{ x\in \RR}\nu^{(x)}(x,y) \quad \text{ s.t. }\quad  (x,y) \in \fC\\
 \min_{y \in \RR}\nu^{(y)}(y,x)  \quad \text{ s.t. }\quad  (x,y) \in \fC\, .
 \end{align*}
The loss function for the $x$-player is $\nu^{(x)}(x,y) =  x^2 - y^2$  and for the $y$-player it is $\nu^{(y)}(y,x) =  y^2 - x^2$.   
 
 We show that this problem is $(1-\epsilon,1)$-strongly benign for any $\epsilon>0$. To avoid repetitive arguments, we only verify all conditions for the $x$-player. Due to the symmetry of the problem, the same argument can be applied to the $y$-player.  
 
 First note that the unique GNE is $[u^{(x)},u^{(y}] = [0,0] \in \fC$
  and the gradient norm $\nabla \nu^{(x)}(0, y) = 0$ for any choice of $ y \in \cX^{(y)}(0)$.  
  Hence, condition \eqref{Benign:uniqueness} of Definition \ref{def:StronglyBenignGNEP} is satisfied. 
  Further, the benign angular condition is satisfied with $\delta=1$ since 
     \[
 \frac{\dprod{\nabla \nu^{(x)}(x,y), x -u^{(x)}}}{\norm{ \nabla \nu^{(x)}(x,y)} \norm{ x -u^{(x)} } } = \frac{\dprod{ 2 x, x  }}{\norm{ 2x} \norm{ x   } }=1    .\] 
 Hence, the benign angular condition in  Definition \ref{def:StronglyBenignGNEP} is satisfied. 
To verify that $D_{\min}^{(x)}(1-\epsilon)>0$, note that $\frac{\nabla \nu^{(x)}(x,y)}{\tnorm{\nabla \nu^{(x)}(x,y)}} = \frac{2x}{\tabsv{2x}}$. Further, $\frac{2x}{\tabsv{2x}}$ is equal to the sign of $x$, that is  $ \frac{2x}{\tabsv{2x}} = -1$ if $x<0$ and $ \frac{2x}{\tabsv{2x}}=1$ for $x>0$ and $ \frac{2x}{\tabsv{2x}}= 0$ for $x =0$.  We denote this by $\sign(x)$. Consider any $[x,y] \in \fC$. We verify that $\relint \cX^{(x)}(y) \ni 0$. Moreover, the distance between $\bd \cX^{(x)}(y)$ and $0$ is always a constant. Together with the observation that $\frac{2x}{\tabsv{2x}} = \sign(x)$, this implies that $D_{\min}(1-\epsilon)>0$ for any $\epsilon>0$. Hence, we conclude that the problem is $(1-\epsilon,1)$-strongly benign. 
 We note that Harker's condition is not satisfied for this example since the constraints are coupled. 
 \end{example}

 \begin{example}[$(\phi, \delta)$-strongly benign GNEP]\label{Example:SB2}
 We consider a two-player GNEP and use the same notation simplification as in Example \ref{Example:SB}. 
 \begin{align*}
 \min_{ x \in \RR^2} \max_{ y \in \RR^2}  
 \norm{ x}^2 - \norm{ y}^2 + \big( x^\top P  y\big)^2
 \quad \text{ s.t. }\quad (x,y) \in \fC. 
 \end{align*}
 Where $P \in \RR^{2\times 2}$ is a positive definite matrix and for $\epsilon>0$
 \begin{align*}
 \fC = \{ (x,y) \in \RR^{2\times 2}: x \in [-1,1]^2, \; y \in[-1,1]^2, x_1 + y_1 + \epsilon x_2 + \epsilon y_2 \leq 1\}.
 \end{align*}
We note that the unique GNE is at $(0,0)$. Similar to the previous example, we can 
check that the conditions for strongly benign GNEP are satisfied.  
 \end{example}

 \subsubsection{Illustrating Examples for the Alternating Coordination \oIPM }
 \begin{example}[Example \ref{Example:SB} continued] In the following we continue with Example \ref{Example:SB}.  Figure \ref{fig:Example1OFPM} shows the execution of the first steps for the online FPM defined in Algorithm~\ref{alg:explore}. The initial iterates are $x_1 = 2, y_1 = 6$ and $\cS_1^{(x)} = [1,3]$, $\cS_1^{(y)} = [5,7]$. We note that $x_1 \in \cS_1^{(x)}$ and $y_1 \in \cS_1^{(y)}$. Further, $\cS_1^{(x)}\times\cS_1^{(y)} \subseteq \fC$. For better visualization, we only plot every second step. The red dot indicates the combined players' iterates, i.e., the vector $[x_t,y_t]$. The boxes denote the product sets $\cS_t^{(x)} \times \cS_t^{(y)}$. The color highlights the algorithm's progress: the first step is in dark red, and the last iterate in dark blue.

\begin{figure}[h]
\center
 	\includegraphics[width=0.8\linewidth]{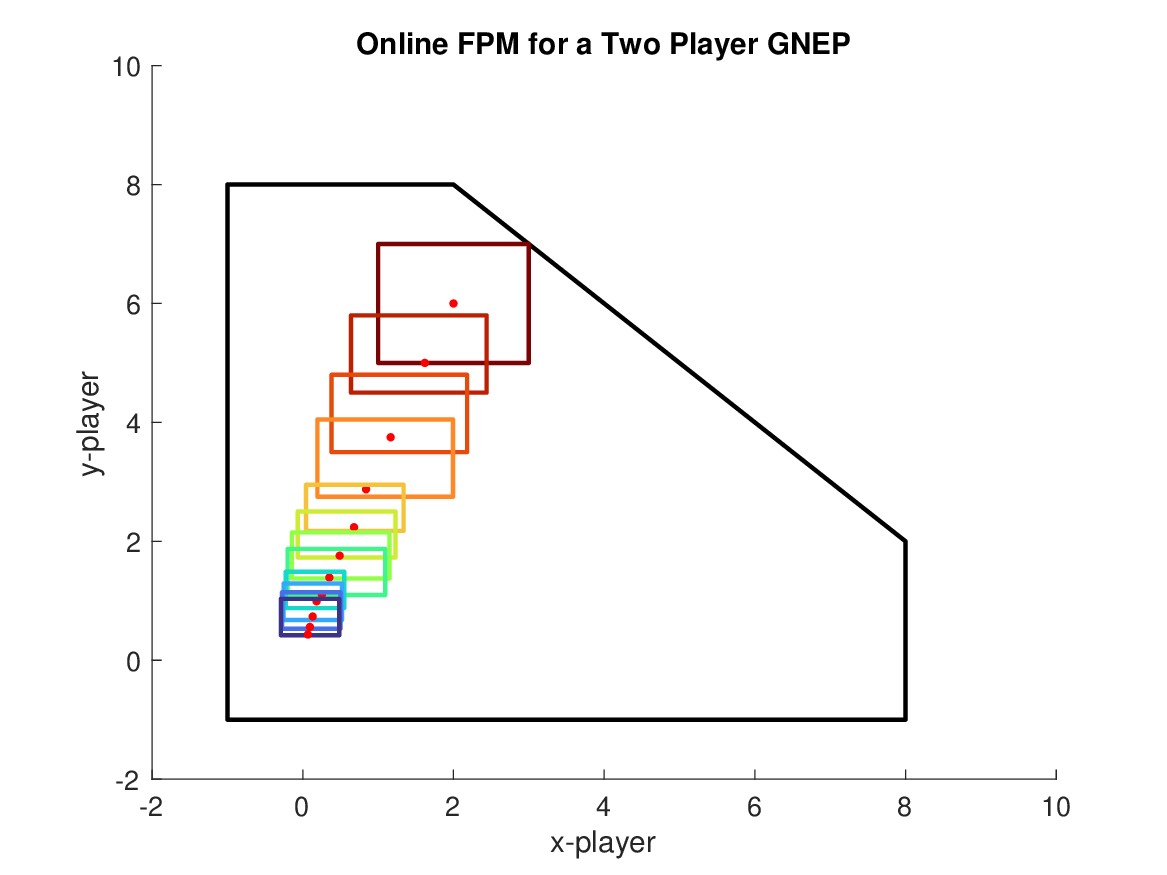}
	 
	\caption{The first 24 iterations of the online FPM Algorithm~\ref{alg:explore} on the GNEP defined in Example~\ref{Example:SB}. 
	}
    \label{fig:Example1OFPM}
\end{figure}

 \end{example}

\subsection{Experiments beyond Benign GNEP}
In this section, we collect further experiments that show that the online FPM converges to GNE beyond its theoretical guarantees. All GNEP in this section are two-player games with $d^{(1)} = d^{(2)} = 1$. As before, we denote the players' actions by $x,y \in \RR$ to eliminate the superscripts.  We only display the iterates for every second step in the plots for the benefit of clarity.

\begin{figure}[t]
\center
 	\includegraphics[width=0.8\linewidth]{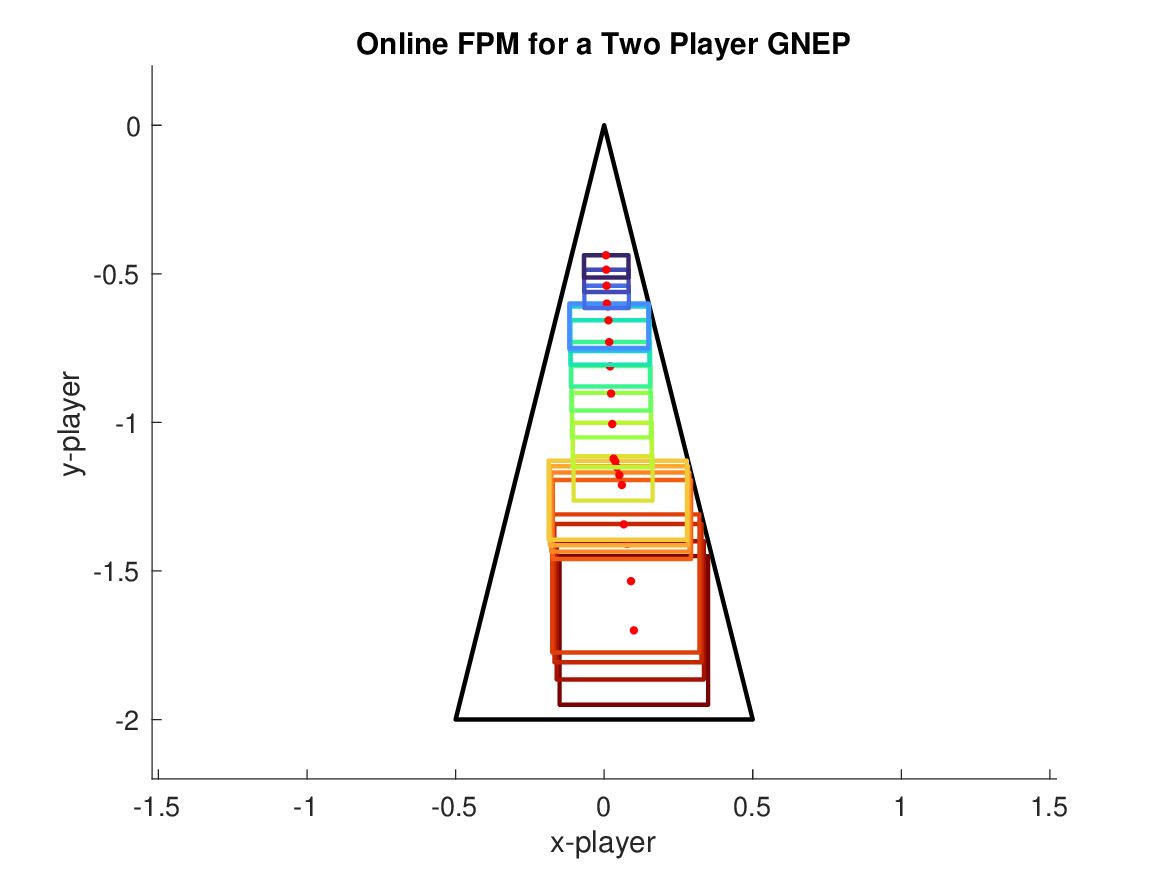}
	 
	\caption{The first 40 iterations of the online FPM Algorithm~\ref{alg:explore} on a GNEP that does not satisfy Condition 1 in Definition \ref{def:StronglyBenignGNEP}. Details in Example~\ref{Example:nB1}. 
	}
    \label{fig:Example:nB1}
\end{figure}

We start with an example that satisfies the benign angular condition and has a unique GNE, but the GNE lies on the boundary. Hence, Condition 1 of Definition \ref{def:StronglyBenignGNEP} is violated.
\begin{example}[Non-benign GNEP]
\label{Example:nB1}
Consider 
\begin{align*}
\min_{x\in\RR} \max_{y \in \RR} x^2 - y^2 \qquad\text{s.t.}\quad [x,y] \in \fC,
\end{align*}
with
\begin{align*}
\fC := \left\{[x,y] \in \RR^2 : y\geq-2,\; 4x + y \leq 0,\;-4x + y \leq 0 \right\}.
\end{align*}
The unique GNE is attained at $[\bar x,\bar y] = [0,0]$. Note that $[\bar x,\bar y] \in \bd \fC$, hence Condition 1 in Definition \ref{def:StronglyBenignGNEP} is not satisfied. However, as can be seen in Figure \ref{fig:Example:nB1}, the online FPM converges to the unique GNE on the boundary.  

\end{example}

 Next, we have an example that does not satisfy Conditions 2 and 3 in Definition \ref{def:StronglyBenignGNEP}: that is,  there does not exist a $\phi$ such that $D_{\min}(\phi)>0$  and the angular condition is violated. However, Condition~1 in Definition \ref{def:StronglyBenignGNEP} is satisfied. 

\begin{figure}[t]
\begin{minipage}[h]{0.3\textwidth}
\includegraphics[scale=0.45]{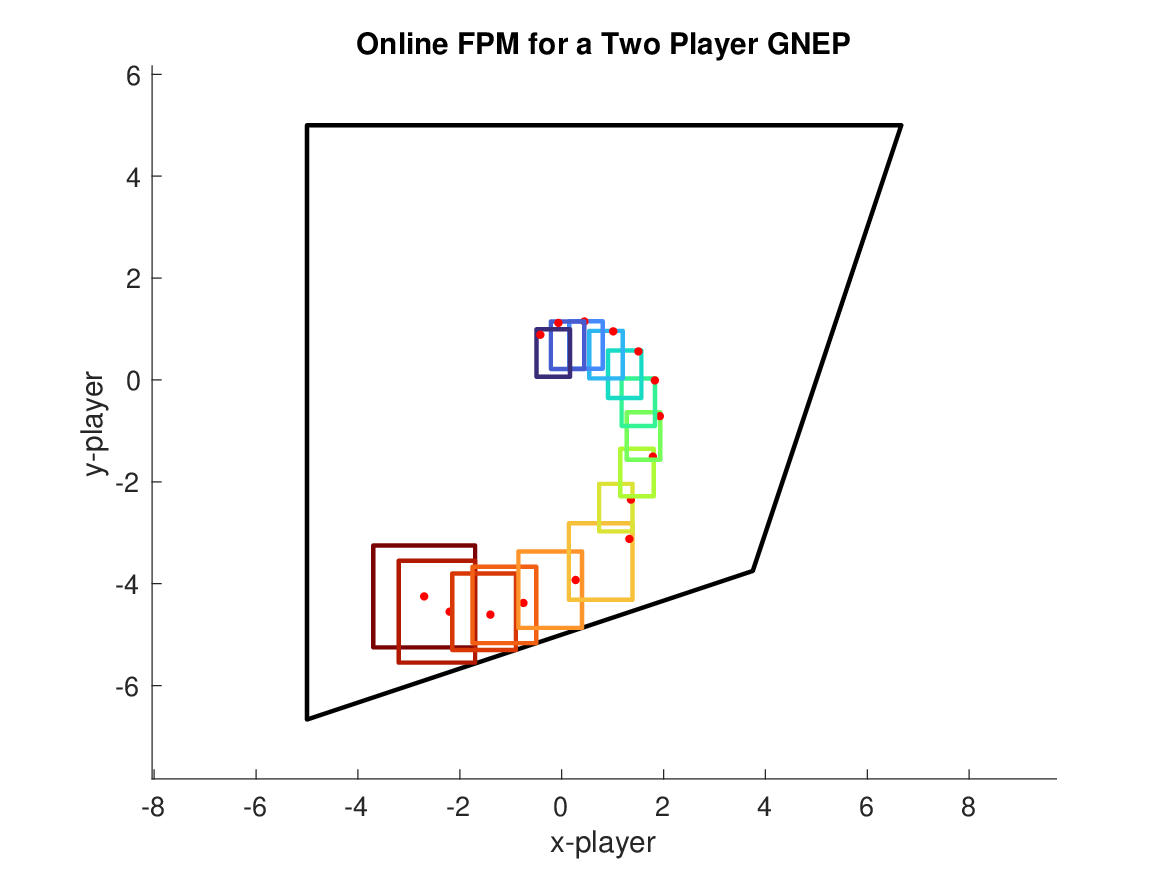}
\end{minipage} \hspace{0.2\textwidth}
\begin{minipage}[h]{0.3\textwidth}
\includegraphics[scale=0.45]{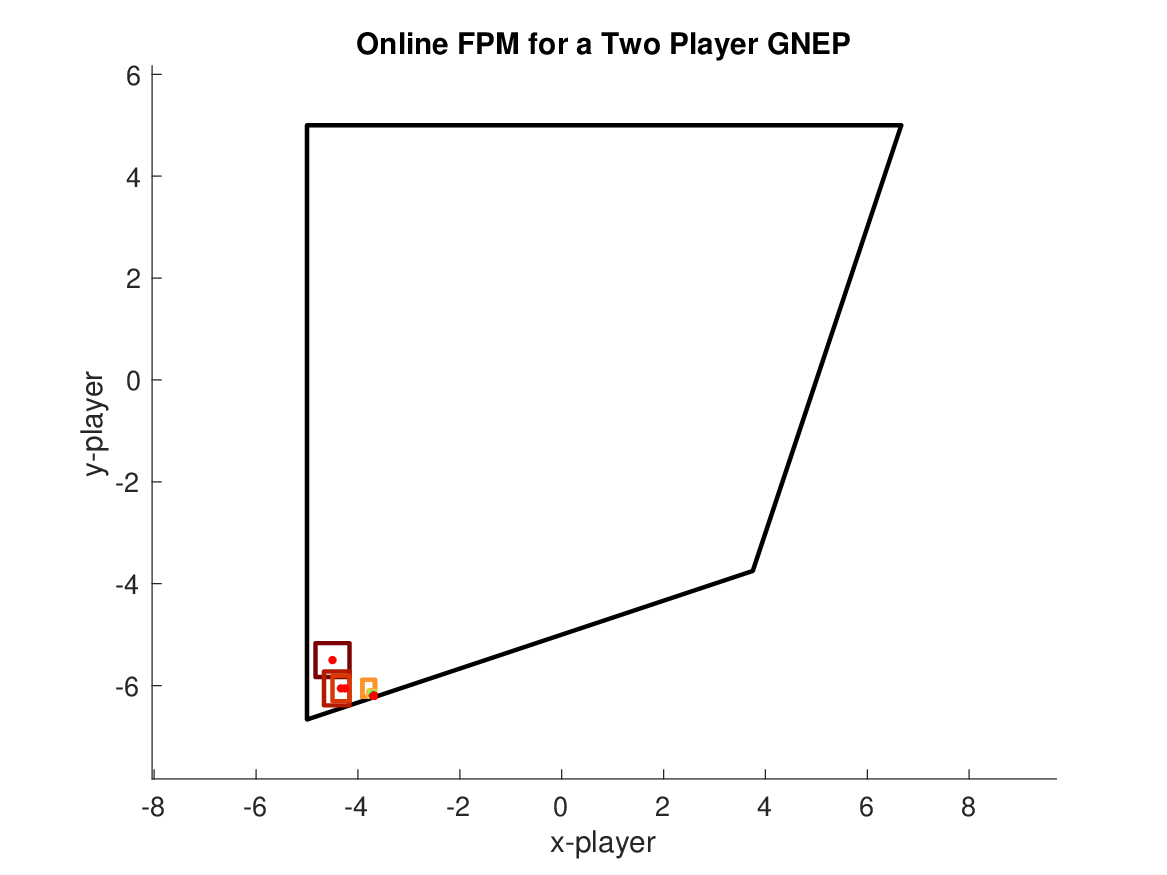}
\end{minipage}
\caption{The first 30 iterations of the online FPM on the GNEP defined in Example \ref{Example:nB2} for two different initializations.}
\label{fig:ExamplenB2}
\end{figure}

\begin{example}[Non-benign GNEP]
\label{Example:nB2}
Consider the following GNEP
\begin{align*}
\min_{x\in\RR} \max_{y \in \RR} x^2 - 10xy - 2y^2 \qquad\text{s.t.}\quad [x,y] \in \fC
\end{align*}
with
\begin{align*}
\fC := \left\{ [x,y] \in \RR^2 : x\geq -5,\; y\leq 5,\; x-\frac{1}{3} y\leq 5,\; y - \frac{1}{3} x \geq- 5 \right\}.
\end{align*}
This problem has a GNE at $[0,0]$. 
As can be seen in Figure \ref{fig:ExamplenB2}, it might converge to the GNE for some initial points. However, initializations exist for which the online FPM converges to a non-equilibrium point as seen in the second plot. 
\end{example}

\section{Feasibility of the Online FPM: Missing Proofs}
\label{proofs:algo}
 
We first show the following propositions.

\begin{restatable}{proposition}{coordination}\emph{[Coordination of Players]}
\label{coordination}
Suppose all players follow Algorithm~\ref{alg:explore}. Then, for all players $i \in [n]$ and all iterations $t \in [T]$, all players are in the same phase $\cP_k$. 
\end{restatable}
\begin{proof}
  This follows directly from the observation that the termination criterion \ref{TC} is the same for all players. Hence, it will be either satisfied for all players or for none. 
 \end{proof}
 \feasibilityExplor*
\begin{proof}
Note that due to the definition of $\iota_t^{(i)}$ and \ref{Vupdate}, $\cS_t^{[\times [n]]} \subseteq \fC$ holds. To show the second result, we show that $x_t^{(i)} \in \relint \cS_t^{(i)}$ for all $i\in [n]$ and all $t \in [T]$. This follows via induction.  More specifically, consider the two cases of the \ref{update}: 
\begin{enumerate}
\item   Iterations with $(t \bmod n) + 1 \neq i$. Note that in this case the $i^{th}$ player has not updated its set, hence $\cS_t^{(i)} = \cS_{t-1}^{(i)}$.  Since $x_{t-1}^{(i)} \in \cS_{t-1}^{(i)}$,  the definition of $\bar \eta_t^{(i)}$ guarantees that $x_t^{(i)} \in \relint \cS_{t}^{(i)}$.  
\item Iterations with $(t \bmod n)  + 1 = i$. In this case, the $i^{th}$ player has translated its set, thus $\cS_t^{(i)} \neq \cS_{t-1}^{(i)}$. Note that due to the definition of \ref{Vupdate} and \ref{update}, the iterate $x_t^{(i)}$ is translated by the same vector as the set $\cS_t^{(i)}$. Since $x_{t-1}^{(i)} \in \cS_{t-1}^{(i)}$, the claim follows. 
\end{enumerate}
 \end{proof}

 \section{Convergence of the Online FPM: Missing Proofs}
\label{proofs:regret}

\subsection{Technical Results}
We first introduce several technical lemmata. We note that most of these results are small variations of existing results. Hence, the following lemmata are primarily included for the convenience of the reader. 
\begin{definition}[Moreau envelope]
Given a function $f: \RR^d \rightarrow \RR$, its Moreau envelope is the function $f^\gamma:\RR^d \rightarrow \RR$ defined by $f^{\gamma}(x) = \inf_{y \in \RR^d} f(y) + \frac{1}{2\gamma}\norm{y-x}^2$.
\end{definition}
Some basic properties of the Moreau envelope.
\begin{enumerate}
\item $f^{\gamma}$ is $(1/\gamma)$-smooth.
\item If $\argmin_{x \in \cX} f(x) $ exists, then $\argmin_{x \in \cX}
f^{\gamma}(x) = \argmin_{x \in \cX} f(x)$.
\end{enumerate}
See, \citet[Proposition 12.29, and 12.30]{Bauschke_2011}.  
We use the Moreau envelope for the norm distance function. For completeness, we add the following derivations: 
\begin{proposition}\label{distMENorm}
Fix $u \in \R^d$ and $a > 0$, and consider the function $f:\RR^d \rightarrow \RR$ defined
by $f(x) =a \norm{x - u}$. Then, for any $\gamma > 0$,
\begin{align*}
f^{\gamma}(x) = \begin{cases}
a\norm{x-u} - \frac{\gamma}{2}a^2 &\text{ if } a\gamma \leq \|x-u\|\\
\frac{ 1}{2\gamma}\norm{x-u}^2 &\text{ if } a\gamma > \|x-u\|
\end{cases},
\end{align*}
and  
$-\frac{\gamma}{2}a^2 \leq f^{\gamma}(x) - f(x)\leq 0$ for all $x \in \RR^d$.
\end{proposition}

\begin{proof}
  By definition of the Moreau envelope,
  \[
     f^{\gamma}(x) = \min_{y \in \R^d}\, \bigg\{ a\norm{y-u} +
     \frac{1}{2\gamma}\norm{y-x}^2 \bigg\} \,.
  \]
  We will use that the minimizing $y$ always lies on the line
  segment $\{(1-\lambda) x + \lambda u : \lambda \in [0,1]\}$ between
  $x$ and $u$. To see this, consider any $y$ not on this line segment.
  Then we can find a $y'$ that does lie on the line segment with smaller
  value, by minimizing the second term $\norm{y'-x}^2$ subject to not
  increasing the first term: $\norm{y'-u} \leq \norm{y-u}$. This $y'$
  will be the projection of $x$ onto a ball around $u$ of radius
  $\norm{y-u}$, and will hence lie on the line segment between $x$ and
  $u$. It follows that
  \begin{align*}
    f^{\gamma}(x)
      &= \min_{\lambda \in [0,1]}\,\bigg\{ a\norm{\big((1-\lambda) x + \lambda
      u\big)-u} + \frac{1}{2\gamma}\norm{\big((1-\lambda) x + \lambda
      u\big)-x}^2 \bigg\}\\
      &= a\norm{x-u} + \min_{\lambda \in [0,1]} \bigg\{
      \frac{\lambda^2}{2\gamma}\norm{x - u}^2 -a\lambda \norm{x-u} \bigg\}.
  \end{align*}
  The minimizer in $\lambda$ of this quadratic is   %
  \[
    \lambda^\star = \begin{cases}
      \frac{a\gamma}{\|x-u\|} & \text{if $a\gamma \leq \|x-u\|$,}\\
      1 & \text{otherwise.}
    \end{cases}
  \]
  Plugging these expressions in, the first part of the proposition follows. For the second part, we observe that the claim is satisfied for any $x$ such that $a\gamma \leq \norm{x-u}$. If $\norm{x-u}<a\gamma$, then $f^\gamma(x) - f(x) + \frac{\gamma}{2}a^2=\frac{1}{2\gamma} \norm{x-u}^2 - a \norm{x-u} + \frac{\gamma}{2}a^2 = \frac{\gamma}{2}(\tnorm{x-u} -a)^2 \geq 0$. In the second case, i.e., whenever $a\gamma>\tnorm{x-u}$, we have $f^\gamma(x) - f(x) = \frac{1}{2\gamma}\tnorm{x-u}^2 - a\norm{x-u} \leq \frac{a\gamma}{2\gamma}\norm{x-u} - a\tnorm{x-u} = -\frac{1}{2}a\tnorm{x-u} \leq 0$.
\end{proof}

 The following lemma is a key technical lemma for our convergence guarantees. We note that the step size requirements are dependent on potentially unknown quantities. However, we show in Corollary \ref{cor:InexactGDSB} and \ref{cor:InexactGD}, that for all cases of interest, the step size requirements are satisfied and can be tuned based on known parameters. 
  \begin{lemma}[Inexact Gradient Descent]
\label{PLInexactGDME}
 Suppose $  \tilde f_t:\RR^d \rightarrow \RR$ is $\tilde
\mu$-strongly convex and differentiable.  
Assume that  there exist $u \in \RR^d$ and $\delta \in (0,1]$ such that for any $t\in[T]$  
  \begin{align}\label{angularCond}
 & \dprod{\nabla\tilde f_t(x_t) , x_t -u
 }  \geq \delta \norm{  \nabla\tilde f_t(x_t) }\norm{ x_t -u}. 
 \tag{\textbf{Angular Condition}}
 \end{align}  
  Consider the update $x_{t+1} = x_t - \eta g_t$ such
that $g_t = \nabla\tilde f_t(x_t)$ and $\eta>0$ denotes the stepsize. Assume $\norm{g_t}\leq G$.  
Consider $f :  \RR^d \rightarrow \RR$   with $f(x) = 2\delta^{-1}   \tnorm{x-u}$. 
   Let $\hat \mu_t \leq \frac{\delta   \norm{g_t}}{\norm{ x_t - u}}$.   Let $C>0$ be any constant. Then for any $t \in \NN$ and $\eta \leq \min(\frac{C}{G\sqrt{T}},  \frac{\delta\tnorm{x_t -u}}{\tnorm{g_t}})$ 

\begin{equation}\label{eqn:approx_grad}
  f(x_t) - f(u) \leq  \Big(f(x_0)-f(u)\Big) \prod_{j=1}^{t-1} \left( 1- \frac{\hat \mu_j
  \eta}{2}\right) + \frac{  C}{\delta^2\sqrt{T} }. 
\end{equation}
\end{lemma}

 \begin{proof}
  Set $\gamma = \frac{C}{2\sqrt{T}}$ and let  $f^{\gamma}$ denote the Moreau envelope with parameter $\gamma$
  of $f$
  (See Proposition \ref{distMENorm}.). Note that due to the basic
  properties of the Moreau envelope $f^{\gamma}$ is $(1/\gamma)$-smooth
  and $f^{\gamma}(u) = f(u) =0$. Further, by Proposition
  \ref{distMENorm},
\begin{align*}
 f(x_t) \leq  f^{\gamma}(x_t) + \frac{\gamma}{2} \frac{4}{\delta^2} = f^{\gamma}(x_t) + \frac{C}{\delta^2\sqrt{T}} .
  \end{align*}
     Because $f^{\gamma}(x_0) \leq f(x_0)$ (again by
  Proposition~\ref{distMENorm}) and
  $f(u)=0$ it is sufficient to show that
\begin{equation}\label{eqn:Moreau_to_show}
  f^{\gamma} (x_{t+1})  \leq f^{\gamma} (x_t)  \Big(1- \frac{\hat
  \mu_t\eta }{2}\Big).
\end{equation}
 The result follows by applying \eqref{eqn:Moreau_to_show} recursively.
Towards proving \eqref{eqn:Moreau_to_show}, we first use
$(1/\gamma)$-smoothness of $f^{\gamma}$ to obtain: 
 \begin{align*}
 f^{\gamma}(x_{t+1}) &\leq  f^{\gamma}(x_t)  + \dprod{\nabla    f^{\gamma}(x_t),
 x_{t+1} - x_t} + \frac{1}{2\gamma}\norm{ x_{t+1 }- x_t}^2\\
&=   f^{\gamma}(x_t) - \eta\dprod{\nabla   f^{\gamma}(x_t), g_t} + \frac{ \eta^2 }{2\gamma}\norm{g_t}^2\\
&= \left(1-\frac{\eta\hat \mu_t}{2}\right) f^{\gamma}(x_t)\underbrace{ - \eta\dprod{\nabla   f^{\gamma}(x_t), g_t} + \frac{ \eta^2 }{2\gamma}\norm{g_t}^2 +\frac{\eta\hat \mu_t}{2} f^{\gamma}(x_t).}_{:= B}
\end{align*}
 Next, we show that $B \leq 0$.
    We distinguish two cases.
  \paragraph{ Case $\norm{x_t - u } \geq 2\frac{\gamma}{\delta}$. }
  In this case, we have $f^{\gamma}(x_t) 
  = 2 \delta^{-1} \norm{x_t - u} - \frac{2\gamma}{\delta^2}
  \leq 2\delta^{-1}\norm{x_t - u} $ by
Proposition~\ref{distMENorm}.
   It follows that $B \leq 0$ is guaranteed when
 \begin{align*}
  &- \eta\dprod{\nabla   f^{\gamma}(x_t), g_t} + \frac{ \eta^2 }{2\gamma}\norm{g_t}^2 +\frac{\eta\hat \mu_t}{2}  2\delta^{-1}\norm{x_t - u}  \leq 0\\
  \Leftrightarrow \quad&\frac{ \eta}{2\gamma}\norm{g_t}^2 \leq \dprod{\nabla   f^{\gamma}(x_t), g_t} -  \hat \mu_t\delta^{-1}\norm{x_t - u}  \,.
 \end{align*}
 We note that $\nabla f^{\gamma} (x_t) = \nabla f(x_t) = \frac{2(x_t-u)}{\delta\norm{x_t-u}}$.   Hence, 
  \[
 \dprod{\nabla f^{\gamma}(x_t), g_t} = \dprod{\nabla f(x_t), g_t} =\frac{2}{\delta\norm{x_t-u}} \dprod{x_t -u,g_t} \overset{(1)}{\geq}
\frac{2}{\delta\norm{x_t-u}}  \delta\norm{g_t}\norm{ x_t -u} = 2  \norm{g_t}  ,
 \]
where we have used the \eqref{angularCond} for inequality (1) (Recall $g_t = \tilde \nabla f_t(x_t)$.). Therefore, to show that
$B\leq 0$, it is suffices to show that
\begin{align*}
  \frac{ \eta}{2\gamma}\norm{g_t}^2 
  \leq  \norm{g_t} + \underbrace{ \norm{g_t}  
  -  \hat \mu_t\delta^{-1}\norm{x_t - u}}_{:= A}
\end{align*}
The term  
$A$ is positive for $\hat \mu_t$ small enough:
\begin{align*}
  \label{ineq:part1}
\norm{g_t} - \hat \mu_t\delta^{-1}\norm{x_t - u} \geq 0 \Leftrightarrow
\frac{\norm{g_t}\delta}{\norm{x_t-u}}\geq\hat \mu_t ,
\end{align*}
which is satisfied due to the assumption that $\hat \mu_t = \frac{\delta   \tnorm{g_t}}{\tnorm{ x_t - u}}$.
 It remains to show that
\[
  \frac{\eta}{2\gamma} \norm{g_t}^2 \leq \norm{g_t}
\]
which is satisfied since by assumption $\eta \leq \frac{2\gamma}{G}$.   Thus, we conclude that $B \leq 0$. 
\paragraph{Case $\norm{x_t - u } < 2\frac{\gamma}{\delta}$. } Then $f^{\gamma}(x_t) = \frac{1}{2\gamma}\tnorm{x_t - u}^2$ (c.f. Proposition~\ref{distMENorm}).   Hence, 
 $B \leq 0$ is guaranteed when
 \begin{align*}
  &- \eta\dprod{\nabla   f^{\gamma}(x_t), g_t} + \frac{ \eta^2 }{2\gamma}\norm{g_t}^2 +\frac{\eta\hat \mu_t}{2}   \frac{1}{2\gamma}\norm{x_t - u}^2  \leq 0\\
  \Leftrightarrow \quad&\frac{ \eta}{2\gamma}\norm{g_t}^2 \leq \dprod{\nabla   f^{\gamma}(x_t), g_t} -   \frac{\eta\hat \mu_t}{4\gamma}  \norm{x_t - u}^2   \,.
 \end{align*}
Further, using that $\nabla f^{\gamma} (x_t) =  \frac{1}{\gamma} (x_t-u)$ gives
  \[
 \dprod{\nabla f^{\gamma}(x_t), g_t} = \frac{1}{\gamma}\dprod{ x_t -u, g_t} \overset{(1)}{\geq}
 \frac{\delta}{\gamma}\norm{g_t}\norm{x_t -u}\, ,
 \]
where inequality (1) is due to the \eqref{angularCond}.  Therefore, to show that
$B\leq 0$, it is suffices to show that
\begin{align*}
  \frac{ \eta}{2\gamma}\norm{g_t}^2 
  \leq   \frac{\delta}{2\gamma}\norm{g_t}\norm{x_t -u} + \underbrace{   \frac{\delta}{2\gamma}\norm{g_t}\norm{x_t -u}
  -    \frac{ \hat \mu}{4\gamma}  \norm{x_t - u}^2}_{:= A}.
\end{align*}
Then, 
\begin{align*}
A\geq 0 \Leftrightarrow     \delta \norm{g_t} 
  -    \frac{ \hat \mu_t}{2}  \norm{x_t - u} \geq 0 \Leftrightarrow
\frac{2\norm{g_t}\delta}{ \norm{x_t-u}}\geq\hat \mu_t ,
\end{align*}
which is satisfied since $\hat \mu_t = \frac{\delta   \tnorm{g_t}}{\tnorm{ x_t - u}}$. It remains to show that
\begin{align}
\label{ineq:stepsizeRequirement2}
 \frac{ \eta}{2\gamma}\norm{g_t}^2 
  \leq   \frac{\delta}{2\gamma}\norm{g_t}\norm{x_t -u} \Leftrightarrow \eta \leq \frac{\delta\norm{x_t -u}}{\norm{g_t}},
\end{align}
which is satisfied by assumption.
This implies $B\leq 0$.

Applying \eqref{eqn:Moreau_to_show} recursively gives the claim. 
\end{proof}

  %%%%%%%%%%%%%%%%%%%%%%%%%%%%%%%%%%%%%%%%
  The step size in Lemma~\ref{PLInexactGDME} seems unusual. However, it reduces to the minimum between $\frac{C}{G\sqrt{T}}$ and $\frac{\delta}{L}$ if the $\tilde f_t$'s are $L$-smooth and $u$ is their common minimizer.  We note that in this case $g_t^u=\nabla \tilde f_t(u) = 0$, hence
   \[\norm{g_t}= \norm{g_t-g_t^u}   \overset{(1)}{\leq} L\norm{x_t-u} . \] 
 Thus, 
 \[
   \frac{\delta\norm{x_t -u}}{\norm{g_t}}\geq   \frac{\delta\norm{x_t -u}}{L\norm{ x_t-u}} = \frac{\delta}{L}.
 \]
 Note that the step size $\frac{\delta}{L}$ differs by the constant $\delta$ from the optimal step size of $\frac{1}{L}$ for deterministic gradient descent and $\frac{C}{G\sqrt{T}}$ reduces to the standard (constant) step size choice for online gradient descent whenever $C = D$. 
 
 Furthermore, we note that for $g_t^u = 0$, setting $\hat \mu = \delta \tilde\mu$  implies that $\hat \mu_t \leq \frac{\delta\tnorm{g_t}}{\tnorm{x_t-u}}$ is satisfied. Indeed, due to strong convexity, we have
 \[ 
 \mu\norm{x_t-u}^2 \leq \dprod{g_t - g_t^u,x_t - u} = \dprod{g_t  ,x_t - u}\leq \norm{g_t}\norm{x_t-u}.
 \]
 Thus, $\norm{x_t-u} \leq \norm{g_t}/\mu$. 
 
 \begin{corollary}
\label{cor:InexactGDSB}
 Suppose all assumptions of Lemma~\ref{PLInexactGDME} are satisfied. Assume that in addition for all $t\in\NN$, $\tilde f_t$ are $L$-smooth and  $u = \argmin_{x \in
\cX } \tilde f_t(x) $ for all $t \in \NN$.  Set $\hat \mu =  \delta \tilde \mu $. Let $C>0$ denote a constant and set $\eta = \min(\frac{C}{G \sqrt{T}},\frac{\delta}{L})$.  Then  
\begin{align*}
  \Big(f(x_t) - f(u)\Big) 
\leq  \Big( 1- q\Big)^{(t-1)}\Big(f(x_0)-f(u)\Big)   
   +   \frac{C}{\delta^2\sqrt{T}}  ,  
 \end{align*}
 where $q = \min\left(\frac{\tilde \mu \delta C}{2G\sqrt{T}}, \frac{\delta^2\tilde \mu}{L}\right)$.
\end{corollary}

The assumption that all $\tilde f_t$'s share a common minimizer is quite strong. Hence, we show a similar corollary under a different assumption: Suppose that for all $t \in \NN$, $\norm{\nabla \tilde f_t(x_t)} \geq \Delta \norm{x_t -u}$. Then we obtain the following corollary. Note that we obtain an additive  $\epsilon$ term in the bound and the step size and convergence rate are dependent on $\epsilon$. 
  \begin{corollary}
\label{cor:InexactGD}
 Suppose all assumptions of Lemma~\ref{PLInexactGDME} are satisfied. Let $u\in\RR^d $ denote the vector for which the \ref{angularCond} of Lemma~\ref{PLInexactGDME} is satisfied. Assume that in addition for all $t\in\NN$, $\tilde f_t$ are $L$-smooth and that there exists a $\Delta>0$ such that $\norm{\nabla \tilde f_t(x_t)} \geq \Delta \norm{x_t -u}$.  Set $\hat \mu =  \Delta \delta $. Let $C>0$ denote a constant and for any $\epsilon>0$ set $\eta_t = \min(\frac{C}{G \sqrt{T}},\frac{\delta \epsilon}{G})$.  Then  
\begin{align*}
  \Big(f(x_t) - f(u)\Big) 
\leq  \Big( 1- q\Big)^{(t-1)}\Big(f(x_0)-f(u)\Big)  
   +   \frac{C}{\delta^2\sqrt{t}}  +2\delta^{-1}\epsilon ,  
 \end{align*}
 where $q =\min\left( \frac{\Delta \delta C}{2G\sqrt{T}}, \frac{\delta^2\Delta \epsilon}{G}\right)$.
\end{corollary}
\begin{proof}
First, note that  $\hat \mu_t =\delta\Delta \leq\frac{\delta \Delta\tnorm{x_t-u}}{\tnorm{x_t-u}}\leq \frac{\tnorm{g_t}\delta}{\tnorm{x_t-u}}$ due to the assumption $\tnorm{\nabla \tilde f_t(x_t)} \geq \Delta \tnorm{x_t -u}$. Hence, $\hat \mu_t \leq \Delta \delta$ implies that the assumptions on $\hat \mu_t$ of Lemma~\ref{PLInexactGDME} are satisfied. Furthermore, we note that for any $x_t$ which is not an $\epsilon$-solution, $\norm{x_t-u}\geq \epsilon$. Thus, $\frac{\delta \norm{x_t-u}}{\norm{g_t}} \geq \frac{\delta\epsilon}{G} \geq \eta$ which implies that the step size assumption of Lemma~\ref{PLInexactGDME} is satisfied. The claim follows by noting that for any $\epsilon$-solution the bound holds. 
\end{proof}

\subsection{Convergence Results}
 We apply the technical lemmata from the previous subsection to show the convergence of the iterates to a GNE. 
  \begin{theorem}[Extended version of Theorem \ref{thm:convergence}]
  \label{thm:convergenceExt}
  Suppose Assumption \ref{CC}  is satisfied and assume all players are following Algorithm \ref{alg:explore} with step size $\eta = \min(\frac{D}{G\sqrt{T}}, \frac{\delta}{L})$. 
 Assume we have a $(\phi,\delta)$-strongly benign GNEP.  Set \[t_0  =  \max\left(   \frac{4D}{\phi} +1,\;  \left(\frac{D}{2D_{\min}}\right)^2, \; \left(\frac{DL}{2G\delta} \right)^2\right).\]  
  Then for all players $i \in [n]$  and any $t  \in[ t_0, T]$ 
    \begin{align*}
  \norm{x_t^{(i)} - u^{(i)}} \leq  \Xi \norm{x_1^{(i)} - u^{(i)}}  \left(1-\frac{ \mu \delta D}{4 G\sqrt{T}}\right)^{  \frac{ t+1}{n}}  +  \frac{2D}{\delta \sqrt{T}} ,
  \end{align*}
  where
   \[
   \Xi =   \left(1-\frac{ \mu \delta D}{4 G\sqrt{T}}\right)^{  -\frac{ t_0}{n}}
       \]
 
   Suppose we have a   
   $(\delta,\phi,\Delta)$-benign GNEP.
   Let $\epsilon>0$ and set  \[t_0  =  \max\left(   \frac{4D}{\phi} +1,\;  \left(\frac{D}{2D_{\min}}\right)^2, \; \left(\frac{D}{2\epsilon\delta} \right)^2\right).\]  
  Then for all players $i \in [n]$  and any $t  \in [t_0, T]$ 
    \begin{align*}
  \norm{x_t^{(i)} - u^{(i)}} \leq  \Xi \norm{x_1^{(i)} - u^{(i)}} \left(1-  \frac{ \delta\Delta D}{2G\sqrt{T}} \right)^{\frac{t+1}{n}}
  +  \frac{1}{\delta}\left(\frac{2D}{ \sqrt{T}}  + 2\epsilon\right) ,
  \end{align*}
  where    
   \[
   \Xi =  \left(1-  \frac{ \delta\Delta D}{2G\sqrt{T}} \right)^{\frac{-t_0}{n}} \]

 \end{theorem}

  \begin{proof}
 We show the convergence result in two steps:
 \begin{enumerate}
 \item We first show that the conditions of Lemma \ref{PLInexactGDME} and Corollary \ref{cor:InexactGDSB} (respectively Corollary \ref{cor:InexactGD} for benign GNEP) are satisfied, and apply it to the loss function of one player. That is, we apply Lemma \ref{PLInexactGDME} with $\tilde f_t(\, \cdot\,) =  \nu(\,\cdot\, , x_t^{(i)})$. 
 This gives a bound of the form  
  \[
 2\delta^{-1}\norm{x_t^{(i)} - u^{(i)}} \leq 2\delta^{-1}\norm{x_1^{(i)} - u^{(i)}} \prod_{s=1}^t(1-\epsilon_s) + \frac{D}{\delta^2\sqrt{T}}.
 \] 
 \item In the second step, for any $t \geq t_0$, we show that$\epsilon_t$ is sufficiently large to guarantee convergence.  
 \end{enumerate}
   We start by showing that all assumptions of Lemma \ref{PLInexactGDME} are satisfied, and then show that the additional assumptions of Corollary \ref{cor:InexactGDSB} and \ref{cor:InexactGD} are satisfied for strongly benign and benign GNEP respectively. 
   First, note that due to  Assumption \ref{CC}, $\nu^{(i)}(\, \cdot\,,x^{-i})$ are $\mu$-strongly convex. 
  Hence, for  $\tilde f_t(\, \cdot\,) =  \nu(\,\cdot\, , x_t^{(i)})$, the  strong convexity assumption of Lemma \ref{PLInexactGDME} is satisfied.\footnote{For the benefit of consistency with respect to the notation used in Assumption \ref{CC}, we keep the notation for the strong convexity parameter of the loss functions $\nu(\,\cdot\, , x_t^{(i)})$ as $\mu$ (not $\tilde \mu$). When applying Lemma \ref{PLInexactGDME}, it corresponds to the strong convexity parameter of the function $\tilde f$.}  
   Next, recall that for  
   benign and strongly benign GNEP the equilibrium is assumed to be unique. Thus, let $(
   u^{(1)}, \dots u^{(n)})$ denote this unique GNE. For this unique GNE,
by definition, the benign angular condition is satisfied (c.f. Definition \ref{def:StronglyBenignGNEP}). That is, there exists a $\delta>0$ such that
   \[ \frac{\dprod{\nabla \utility{i}{}{x}{x},x^{(i)} -u^{(i)}}}{\norm{\nabla \utility{i}{}{x}{x}} \norm{x^{(i)} - u^{(i)}} } \geq \delta.\]
   Thus,  the \ref{angularCond} of Lemma \ref{PLInexactGDME} is satisfied.  
   Next, recall that   $\eta =  \min(\frac{D}{G\sqrt{T}}, \frac{\delta}{L})$ and by the definition of the \ref{update} step, a player either takes a gradient descent step with step size $ \min(  \eta, \bar\eta_t^{(i) }/2 )$ or $ \min(  \eta, \iota_t^{(i)} )$.  In both cases  the step size is bounded by $  \min(\frac{D}{G\sqrt{T}}, \frac{\delta}{L})$. Setting $\gamma = \frac{D}{2\sqrt{T}}$, this implies that for strongly benign GNEP, the step size conditions of Corollary \ref{cor:InexactGDSB} are satisfied and consequently also the step size requirements of Lemma \ref{PLInexactGDME}.  
  Further, for a strongly benign GNEP, $\norm{\nabla \utility{i}{}{u}{x}} = 0$. Thus, the assumptions for Corollary \ref{cor:InexactGDSB} are satisfied.  Setting $\hat \mu_t = \frac{ \mu \delta}{2}$ for all $t \in \NN$ gives
 for any $k \in \NN$ and $t \in \cP_k$  
 \begin{align}
 \label{ineq:boundSB}
 2\delta^{-1} \norm{x_t^{(i)} - u^{(i)}} \leq  2\delta^{-1}\norm{x_1^{(i)} - u^{(i)}} \prod_{s =1}^t\left( 1- \frac{ \mu \delta   \epsilon^{(i)}_s}{4}\right) +  \frac{D}{\delta^2 \sqrt{T}} ,
 \end{align}
 where $\epsilon_s^{(i)} =  \min(  \eta, \bar\eta_t^{(i) }/2 )$ if $s \bmod n \neq i-1$ and $\epsilon_s^{(i)} =  \min(  \eta, \iota_t^{(i)} )$ otherwise. 
 This establishes the first step of the proof.

Next, we show that some of the $\epsilon_t^{(i)}$'s are sufficiently large, i.e., in the order of $ 1/\sqrt{T}$. We emphasize that this cannot be shown for every iteration, but only for the iteration where $t \bmod n = i-1$ and $t \geq t_0$. Recall that these are the rounds where the $i^{\mathrm{th}}$ player shifts its set $\cS_t^{(i)}$. We proceed by inferring that for these iterations, the player's step size can be guaranteed to be in the order of $ 1/\sqrt{T}$. 

 Consider any $t\geq t_0$, $i \in [n]$ and $x_t^{(i)} \in \cS_t^{(i)}$. We denote $g_t^{(i)} =  \nabla \utility{i}{t}{x}{x} $. Then, due to the definition of $D_{\min}(\phi)$ 
\begin{align*}
\hat x_{t+1}^{(i)}: = x_{t}^{(i) } - \frac{D_{\min}(\phi)}{\norm{g_t^{(i)}}} g_t^{(i)} \in \cX^{(i)}(x^{(-i)}).   
 \end{align*}
 Since the set $ \cX^{(i)}(x^{(-i)})$ is convex, and since $x_t^{(i)} \in  \cX^{(i)}(x^{(-i)})$, we know that for any stepsize  $\eta \leq  D_{\min}(\phi)/\norm{g_t^{(i)}}$, $x_t^{(i)} - \eta g_t^{(i)} \in   \cX^{(i)}(x^{(-i)})$.
  Due to $t_0 \geq (D/(2D_{\min}(\phi)))^2$, we know that for any $t \geq t_0$, $\eta = \min(\frac{\delta}{L}, \frac{D}{G\sqrt{T}})\leq \frac{D}{G\sqrt{t}} \leq \frac{D_{\min}(\phi)}{2G}$. Hence, $x_{t+1}^{(i)} = x_t^{(i)} - \eta g_t^{(i)} \in \cX^{(i)}(x^{(-i)})$.
  It remains to show that $\cS_t^{(i)} - \{ \eta g_t^{(i)} \}
  \subseteq \cX^{(i)}(x^{(-i)})$. For this, note that 
   \[
    t_0 \geq \frac{4D}{\phi} +1 \geq \sum_{j=1}^{\ceil{\log_2\frac{D}{\phi}}} 2^j    \, .
  \] 
 Set $k_0 =  \ceil{\log_2  (D/\phi)}$ and recall that due to \ref{TC}, the diameters of the sets $\cS_t^{(i)}$ are decreased by $\frac{1}{2}$ at least every $2^k$ iterations.  Thus, for any $t \geq t_0$, \[\Diameter{\cS_t^{(i)}} \leq \frac{D}{2^{k_0}} \leq \phi,\]  and due to the definition of $D_{\min}(\phi)$, $\cS_t^{(i)} - \{ \eta g_t^{(i)} \} \subseteq \cX^{(i)}(x^{(-i)})$. This gives that for any $t \geq t_0$
  \begin{align*}
 2\delta^{-1} \norm{x_t^{(i)} - u^{(i)}} &\leq  2\delta^{-1}\norm{x_1^{(i)} - u^{(i)}} \prod_{s =t_0}^t\left( 1- \frac{\hat \mu_s   \eta}{4}\right) +   \frac{D}{\delta^2 \sqrt{T}}\\
   &\overset{(1)}{\leq}  2\delta^{-1}\norm{x_1^{(i)} - u^{(i)}} \prod_{\substack{s =t_0\\ (s \bmod n) +1 = i}}^t\left( 1- \frac{ \mu  \delta D}{4G\sqrt{T}}\right) +  \frac{D}{\delta^2 \sqrt{T}} \\
   &\leq 2\delta^{-1} \norm{x_1^{(i)} - u^{(i)}}  \left(1-\frac{ \mu \delta D}{4 G\sqrt{T}}\right)^{  \frac{t+1 - t_0}{n}}+  \frac{D}{\delta^2 \sqrt{T}} \\
   &=  \Xi  2\delta^{-1}\norm{x_1^{(i)} - u^{(i)}}   \left(1-\frac{ \mu \delta D}{4 G\sqrt{T}}\right)^{  \frac{ t+1}{n}}+   \frac{D}{\delta^2 \sqrt{T}} .
 \end{align*}
 Where inequality $(1)$ follows since for any $t\geq t_0$ the minimum for the step size definition $\min\left(\frac{D}{G\sqrt{T}}, \frac{\delta}{L} \right)$ is attained at $\frac{D}{G\sqrt{T}}$. Further, we used here that $\hat \mu_s = \frac{\mu \delta}{2}$.

 For the second part of the result, we note that the conditions of Corollary \ref{cor:InexactGD} are satisfied. Indeed, the only difference is the assumption that gradient norms satisfy $\norm{\nabla \utility{i}{}{u}{x}} \geq \Delta\norm{x_t^{(i)} - u^{(i)}}$, which is satisfied due to the assumptions for benign GNEP. Set $\hat \mu =  \delta\Delta$. Then analogously to \eqref{ineq:boundSB},
  \begin{align}
2\delta^{-1}  \norm{x_t^{(i)} - u^{(i)}} \leq 2\delta^{-1}\norm{x_1^{(i)} - u^{(i)}} \prod_{s =1}^t\left( 1- \frac{  \delta \Delta \eta}{2 }\right) +  \frac{4GD}{\delta^2 \sqrt{T}} + 2\frac{\epsilon}{\delta}.
 \end{align}
 Noting that the rest of the argument is analogous and differs only with respect to the constants, gives the second result. 
\end{proof}

\section{Relations and Examples for Strongly Benign GNEP: Missing Proofs}
\label{appendix:SB}
\RelationSB*
\begin{proof}
\begin{enumerate}
\item  Due to $\mu$-strong convexity and $L$-smoothness, we have that for all players $i \in [n]$
\begin{align*}
 &\dprod{\nabla \utility{i}{}{x}{x} - \nabla \utility{i}{}{u}{x},x^{(i)} -u^{(i)}}\geq  \mu \norm{x^{(i)} - u^{(i)}}^2\\
  &\phantom{xxxxxxxxxxxxxxxx} \geq   \mu L \norm{x^{(i)} - u^{(i)}}\norm{\nabla \utility{i}{}{x}{x}- \nabla \utility{i}{}{u}{x}}.
\end{align*}
  Using that  $\nabla \utility{i}{}{u}{x} = 0$ by assumption and reordering the terms gives the benign angular condition with parameter $\mu L$. 
\item Due to the benign angular condition, we have 
\begin{align*}
 \dprod{\nabla \utility{i}{}{x}{x},x^{(i)} -u^{(i)}}  \geq \delta \norm{\nabla \utility{i}{}{x}{x}} \norm{x^{(i)} - u^{(i)}}.
\end{align*}
 We note that for strongly benign GNEP there exists a unique GNE $u \in  \fC$ such that for all $x \in \fC$, we have $\nabla \utility{i}{}{u}{x} = 0$. Thus, $\tnorm{\nabla \utility{i}{}{x}{x} } = \tnorm{\nabla \utility{i}{}{x}{x}-\nabla\utility{i}{}{u}{x}} $ and due to $L$-bi-Lipschitzness $\tnorm{\nabla \utility{i}{}{x}{x}-\nabla\utility{i}{}{u}{x}} \geq L^{-1}\tnorm{x^{(i)} - u^{(i)}}$. Overall, we obtain
 \begin{align*}
 \dprod{\nabla \utility{i}{}{x}{x}-\nabla\utility{i}{}{u}{x},x^{(i)} -u^{(i)}}  \geq \frac{\delta}{ L}   \norm{x^{(i)} - u^{(i)}}^2.
 \end{align*}
 Summing over all players implies $\frac{\delta}{L}$-strong monotonicity. 
\end{enumerate}
\end{proof}

  \section{Regret Bounds}
  \label{appendix:regretBounds}
  \regretBoundSB*
  
    \begin{proof}
  Due to convexity and since by assumption $\tnorm{\nabla  \nu^{(i)}(x_t^{(i)},x_t^{(-i)})}\leq G$, we have
  \begin{align*}
 \nu^{(i)}\left(x_t^{(i)},x_t^{(-i)}\right)  -   \nu^{(i)}\left(u^{(i)},x_t^{(-i)}\right) &\leq \dprod{\nabla  \nu^{(i)}\left(x_t^{(i)},x_t^{(-i)}\right), x_t^{(i)} - u^{(i)}}\\
  &\leq \norm{\nabla  \nu^{(i)}\left(x_t^{(i)},x_t^{(-i)}\right)}\norm{x_t^{(i)} - u^{(i)}}\\ &\leq G\norm{x_t^{(i)} - u^{(i)}}.
  \end{align*}
  Summing over $t_0$ to $T$ rounds, we obtain
  \begin{align*}
  \sum_{t=t_0}^T \nu^{(i)}\left(x_t^{(i)},x_t^{(-i)}\right)  -   \nu^{(i)}\left(u^{(i)},x_t^{(-i)}\right)  &\leq 
  \frac{\delta G}{2}\sum_{t=t_0}^T2\delta^{-1}\norm{x_t^{(i)} - u^{(i)}}.
  \end{align*}
  Next, we apply Theorem~\ref{thm:convergence} to all terms with $t \geq t_0$. 
  This gives
  \begin{align*}
  \sum_{t=t_0}^T \nu^{(i)}\left(x_t^{(i)},x_t^{(-i)}\right)  -   &\nu^{(i)}\left(u^{(i)},x_t^{(-i)}\right) \leq \frac{\delta G}{2}\left(D\Xi \sum_{t=t_0+1}^T \left(1-\frac{\delta \mu D}{4G\sqrt{T}} \right)^{\frac{t+1}{n}} + \frac{2D\sqrt{T}}{\delta}\right)\\
&\leq \frac{\delta GD\Xi}{2}\int_{t_0}^T \left(1-\frac{\delta \mu D}{4G\sqrt{T}} \right)^{\frac{x+1}{n}} dx+ DG\sqrt{T} \\
&= \frac{\delta GD\Xi}{2} \left( n \frac{\left(\left( 1-\frac{\delta \mu D}{4G\sqrt{T}} \right)^{\frac{t_0+1}{n}}- \left( 1-\frac{\delta \mu D}{4G\sqrt{T}} \right)^{\frac{T+1}{n}}\right)}{-\log\left( 1-\frac{\delta \mu D}{4G\sqrt{T}}  \right)}\right)+ DG\sqrt{T}\\
&\overset{(1)}{\leq}  \frac{\delta GD\Xi}{2} \left( n \frac{4G\sqrt{T}}{\delta \mu D}\right)+ DG\sqrt{T}\, .
  \end{align*}
  Where we applied that for all $x>-1$ it holds that $ \log(1+x) \leq x$. Furthermore, recall that $t_0$ is a constant independent of $T$. Hence, we bound the first $t_0$ rounds using convexity and the Cauchy-Schwarz inequality: 
  \begin{align*}
   \sum_{t=1}^{t_0} \nu^{(i)}\left(x_t^{(i)},x_t^{(-i)}\right)  -   \nu^{(i)}\left(u^{(i)},x_t^{(-i)}\right)  &\leq 
   \sum_{t=1}^{t_0}  \dprod{\nabla  \nu^{(i)}\left(x_t^{(i)},x_t^{(-i)}\right), x_t^{(i)} - u^{(i)}}\\
  &\leq   \sum_{t=1}^{t_0} \norm{\nabla  \nu^{(i)}\left(x_t^{(i)},x_t^{(-i)}\right)}\norm{x_t^{(i)} - u^{(i)}}\\ 
  &\leq t_0 G D, 
  \end{align*}
  which shows the claim. 
  \end{proof}

 \subsection{Sublinear Regret for Constrained Online Convex Optimization}
 \label{COCO}

\newcommand{\diam}{\Diameter}
 
\begin{lemma}
\label{regretSinglePlayer}
  Let $S_t \subset \RR^d$, $t=1,2,\ldots$ be closed, non-empty, convex
  sets of diameter at most $\diam{S_t} \leq D$ and such that the
  maximum distances between consecutive sets are small:
  \[
    \max_{a \in S_t} \dist(S_{t+1},a) \leq \omega_{t+1}
  \]
  for some $\omega_2,\omega_3,\ldots$ Consider online gradient descent
  with $x_1 \in S_1$ and $x_{t+1} = \proj{S_{t+1}}{x_t - \eta_t g_t}$.
  If the gradients $g_t = \nabla f_t(x_t)$ are uniformly bounded by
  $\|g_t\| \leq G$, the functions $f_t : \RR^d \to \RR$ are convex and
  the step-sizes $\eta_1 \geq \cdots \geq \eta_T > 0$ are
  non-increasing. Then, for any $u \in \RR^d$,
  \[
    \Regf(u) 
    \leq \frac{1}{2\eta_T} (D +
    \max_{t \leq T} \dist(S_t,u))^2
      + G^2 \sum_{t=1}^T \frac{\eta_t}{2}
      + \sum_{t=1}^{T-1} \big(\frac{\omega_{t+1}}{\eta_t} +
      G\big)\dist(S_{t+1},u) -  \frac{1}{2 \eta_T} \|x_{T+1} - u\|^2.
  \]
  In particular, if $\dist(S_t,u) \leq \frac{c}{\sqrt{t}}$, $\eta_t =
  \frac{D + c}{G \sqrt{t}}$ and $\omega_{t+1} \leq c'G\eta_t$, then
  \[
    \Regf(u)
      \leq \frac{3D + (6+4c')c}{2}G\sqrt{T}.
  \]
  If the functions $f_t : \RR^d \to \RR$ are also $\mu$-strongly convex
  and the step-sizes $\eta_t =  \frac{1}{t\mu} $, then, for any $u \in \RR^d$,
  \[
    \Regf(u) 
    \leq  \frac{G^2}{\mu}\log T  
      + \sum_{t=1}^{T-1} \big(\frac{\omega_{t+1}}{\eta_t} + G\big) \dist(S_{t+1},u)
      -  \frac{1}{2 \eta_T} \|x_{T+1} - u\|^2.
  \]

  \end{lemma}

\begin{proof}
  The proof follows the standard OGD analysis, except that we need to be
  more careful when applying the Pythagorean inequality, because it may
  be the case that $u \not \in S_t$. Let $\tilde x_{t+1} = x_t - \eta_t
  g_t$ be the unprojected update, and let $u_t = \proj{S_t}{u}$. Then
  \begin{align*}
    \|\tilde x_{t+1} - u\|^2
    &=
    \|\tilde x_{t+1} - u_{t+1} + u_{t+1} - u\|^2\\
    &=
    \|\tilde x_{t+1} - u_{t+1}\|^2 + \|u_{t+1} - u\|^2 + 2\dprod{\tilde
    x_{t+1} - u_{t+1},u_{t+1} - u}\\
    &\geq
    \|\tilde x_{t+1} - x_{t+1}\|^2 + \|x_{t+1} - u_{t+1}\|^2 + \|u_{t+1} - u\|^2 + 2\dprod{\tilde
    x_{t+1} - u_{t+1},u_{t+1} - u}\\
    &=
    \|\tilde x_{t+1} - x_{t+1}\|^2 + \|x_{t+1} - u\|^2
      - 2\dprod{x_{t+1} - u_{t+1},u_{t+1} - u}
      + 2\dprod{\tilde x_{t+1} - u_{t+1},u_{t+1} - u}\\
    &=
    \|\tilde x_{t+1} - x_{t+1}\|^2 + \|x_{t+1} - u\|^2
      + 2\dprod{\tilde x_{t+1} - x_{t+1},u_{t+1} - u}\\
    &\geq
    \|\tilde x_{t+1} - x_{t+1}\|^2 + \|x_{t+1} - u\|^2
      - 2\|\tilde x_{t+1} - x_{t+1}\| \|u_{t+1} - u\|\\
    &\geq
    \|x_{t+1} - u\|^2 - 2\|\tilde x_{t+1} - x_{t+1}\| \|u_{t+1} - u\|,
  \end{align*}
  where the first inequality follows by the Pythagorean inequality, and the
  second one by Cauchy-Schwarz.
  Observing that
  \begin{align*}
    \|\tilde x_{t+1} - x_{t+1}\|
    &\leq \|\tilde x_{t+1} - \proj{S_{t+1}}{x_t}\|
    = \|x_t - \proj{S_{t+1}}{x_t} - \eta_t g_t\|\\
    &\leq \|x_t - \proj{S_{t+1}}{x_t}\| + \eta_t \|g_t\|
    \leq \omega_{t+1} + \eta_t \|g_t\|,
  \end{align*}
  we conclude that
  \[
    \|\tilde x_{t+1} - u\|^2
    \geq \|x_{t+1} - u\|^2 - 2(\omega_{t+1} + \eta_t \|g_t\|)
    \dist(S_{t+1},u).
  \]
  It follows that
  \begin{align*}
  \sum_{t=1}^T f_t(x_t) - f_t(u)
    &\leq \sum_{t=1}^T \dprod{x_t-u,g_t}\\
    &= \sum_{t=1}^T \frac{1}{2 \eta_t} \Big(\|x_t - u\|^2 - \|\tilde x_{t+1} -
      u\|^2\Big) + \sum_{t=1}^T \frac{\eta_t}{2} \|g_t\|^2\\
    &\leq \sum_{t=1}^T \frac{1}{2 \eta_t} \Big(\|x_t - u\|^2 - 
      \|x_{t+1} - u\|^2
      + 2(\omega_{t+1} + \eta_t \|g_t\|) \dist(S_{t+1},u)
    \Big) + \sum_{t=1}^T \frac{\eta_t}{2} \|g_t\|^2\\
    &= \sum_{t=1}^T \frac{1}{2 \eta_t} \Big(\|x_t - u\|^2 - 
      \|x_{t+1} - u\|^2 \Big) 
      + \sum_{t=1}^T \big(\frac{\omega_{t+1}}{\eta_t} + \|g_t\|\big) \dist(S_{t+1},u)
      + \sum_{t=1}^T \frac{\eta_t}{2} \|g_t\|^2\\
    &= 
      \frac{1}{2 \eta_1} \|x_1 - u\|^2 - \frac{1}{2 \eta_T} \|x_{T+1} - u\|^2
      + \sum_{t=2}^T \Big(\frac{1}{2 \eta_t} - \frac{1}{2
      \eta_{t-1}}\Big) \|x_t - u\|^2\\
    &\quad+ \sum_{t=1}^T \big(\frac{\omega_{t+1}}{\eta_t} + \|g_t\|\big)\dist(S_{t+1},u)
      + \sum_{t=1}^T \frac{\eta_t}{2} \|g_t\|^2\\
    &\leq \frac{(D+\max_{t\leq T} \dist(S_t,u))^2}{2 \eta_T} -  \frac{1}{2 \eta_T} \|x_{T+1} - u\|^2\\
     &\qquad+ \sum_{t=1}^T \big(\frac{\omega_{t+1}}{\eta_t} + G\big)\dist(S_{t+1},u)
      + G^2 \sum_{t=1}^T \frac{\eta_t}{2}  .
  \end{align*}
  Since $S_{T+1}$ is not used in the algorithm, we assume that $S_{T+1}
  = \RR^d$ without loss of generality, so that $\dist(S_{T+1},u) = 0$.
  The first result of the theorem then follows.

  For the second result we plug in the extra assumptions, which gives
  \begin{align*}
    \Regf(u)
      \leq \frac{(D + c)G}{2}\sqrt{T} + \frac{(D+(3+2c')c)G}{2} \sum_{t=1}^T
      \frac{1}{\sqrt{t}}.
  \end{align*}
  Using that $\sum_{t=1}^T \frac{1}{\sqrt{t}} \leq 2\sqrt{T}-1 \leq
  2\sqrt{T}$, the second result then follows.

  Next, we show the regret bound for strongly convex functions. As usual, we use that $\frac{1}{2\eta_t} - \frac{\mu}{2} = \frac{1}{2\eta_{t-1}}$. Note that for $t=1$, this is zero.  
   \begin{align*}
  \sum_{t=1}^T &f_t(x_t) - f_t(u)
    \leq \sum_{t=1}^T \dprod{x_t-u,g_t} - \frac{\mu}{2}\norm{x_t - u}^2\\
    &\leq \sum_{t=1}^T \Big(\left( \frac{1}{2 \eta_t} - \frac{\mu}{2}\right) \|x_t - u\|^2 - 
     \frac{1}{2\eta_t} \|x_{t+1} - u\|^2\Big)\\
      &\quad+ \sum_{t=1}^T \big(\frac{\omega_{t+1}}{\eta_t} + \|g_t\|\big) \dist(S_{t+1},u) 
     + \sum_{t=1}^T \frac{\eta_t}{2} \|g_t\|^2\\
     &= \sum_{t=2}^T  \frac{1}{2 \eta_{t-1}}   \|x_t - u\|^2 - \sum_{t=1}^T
     \frac{1}{2\eta_t} \|x_{t+1} - u\|^2\\
      &\quad+ \sum_{t=1}^T \big(\frac{\omega_{t+1}}{\eta_t} + \|g_t\|\big) \dist(S_{t+1},u)
     + \sum_{t=1}^T \frac{\eta_t}{2} \|g_t\|^2\\
     &= -\frac{1}{2\eta_T} \|x_{T+1} - u\|^2
      + \sum_{t=1}^T \big(\frac{\omega_{t+1}}{\eta_t} + \|g_t\|\big) \dist(S_{t+1},u)
     + \sum_{t=1}^T \frac{\eta_t}{2} \|g_t\|^2.
       \end{align*}
       Together with the bound $\sum_{t=1}^T
       \frac{\eta_t}{2}\norm{g_t}^2 \leq \frac{G^2}{\mu} \log T$ and
       assuming again without loss of generality that $S_{T+1} =
       \RR^d$,
       the last result follows. 
\end{proof}

 \section{Example with linear constraints}

%\todos{This is nice, but how should we integrate it in the main part?}

Consider the problem 
\[
  f(x, y) = \frac{\|x\|^2}{2} -\frac{\|y\|^2}{2} + \langle x, Ay \rangle
\]
with no constraints, this problem admits a unique equilibrium at $(0, 0)$. 

Consider the single affine constraint for some $B > 0$
\[
  \langle x, c_x \rangle + \langle y, c_y \rangle \leq B
  \,. 
\]
Since $B> 0$, the unconstrained equilibrium $(0, 0)$ remains an equilibrium of
the constrained problem. Let us compute the other generalized equilibria of this
problem if there are any. 

A pair of points $(x^\star, y^\star)$ is an equilibrium if and only if
\[
  x^\star \in \argmin \bigg\{ \frac{\|x\|^2}{2} + \langle x, Ay \rangle \bigg\}
  \quad \text{u.c.} \quad
  \langle x, c_x \rangle + \langle y, c_y \rangle \leq B
  \,.
\]
and 
\[
  y^\star \in \argmin \bigg\{ \frac{\|y\|^2}{2} - \langle x, Ay \rangle \bigg\}  
  \quad \text{u.c.} \quad
  \langle x, c_x \rangle + \langle y, c_y \rangle \leq B
  \,.
\]

Let $(x, y)$ be an equilibrium. If 
\[
  \langle x, c_x \rangle + \langle y, c_y \rangle < B
\]
then 
\[
    (x, y)  = (0, 0) . 
\]

Otherwise, there exists Lagrange multipliers $\alpha \geq 0$ and $\beta \geq 0$, such that 
\begin{align*}
&x + Ay = - \alpha c_x \\
&y - A^\top x = - \beta c_y \\
&\langle x, c_x \rangle + \langle y, c_y \rangle = B 
\end{align*}
Therefore, solving the first two equations for $x$ and $y$, 
\begin{align}
\label{eq:solution_equilibrium_x}
&x = - \alpha (I + AA^\top)^{-1}c_x + \beta(I + AA^\top)^{-1} A c_y \\
\label{eq:solution_equilibrium_y}
&y = - \alpha (I + A^\top A)^{-1}A^\top c_x - \beta(I + A^\top A)^{-1} c_y \\
&\langle x, c_x \rangle + \langle y, c_y \rangle = B
\end{align}
And in particular,
\begin{multline*}
B = \langle x, c_x \rangle + \langle y, c_y \rangle \\
= - \alpha \underbrace{\langle 
  c_x, (I + AA^\top)^{-1} c_x + A(I + A^\top A)^{-1} c_y 
  \rangle}_{:= - u_1}
 - \beta \underbrace{
   \langle c_y, (I + A^\top A)^{-1} c_y - A^\top(I +  A A^\top)^{-1} c_x 
  \rangle}_{:= - u_2}
\end{multline*}
To summarize, if $x, y$ is an equilibrium on the border of the constraint set, 
then there exists $\alpha, \beta \geq 0$ such that 
\[
B = \alpha u_1 + \beta u_2. 
\]
Conversely, whenever such a pair exists, one can check that the solutions
\eqref{eq:solution_equilibrium_x} and \eqref{eq:solution_equilibrium_y} form an
equilibrium.
 Since $B$ is positive, the set of values of $\alpha, \beta$ that satisfy this is
either 
\begin{itemize}
  \item empty if $u_1 < 0$ and $u_2 < 0$, 
  \item a segment in $\R^2$ if $u_1 \geq 0$ and $u_2 \geq 0$, 
  \item a half-line in $\R^2$ otherwise. 
\end{itemize}
Note however that
\[
-\langle c_y, (I + A^\top A)^{-1} c_y \rangle
- \langle c_x, (I +  AA^\top)^{-1} c_x \rangle
- \langle c_x,  A(I + A^\top A)^{-1} c_y \rangle
+ \langle  c_y,  A^\top(I + AA^\top)^{-1} c_x \rangle
\leq 0 
\]
since $(I + A^\top A)$ and  $(I + AA^\top)$ are definite positive and
\begin{multline*}
  \langle c_x,  A(I + A^\top A)^{-1} c_y \rangle
  - \langle  c_y,  A^\top(I + AA^\top)^{-1} c_x \rangle
  = \langle c_x, (A(I + A^\top A)^{-1}  - (I + AA^\top)^{-1}A)c_y\rangle \\
  = \langle c_x, ((A^{-1} + A^\top)^{-1}  - (A^{-1} + A^\top)^{-1}A)c_y\rangle
  = 0 \,. 
\end{multline*}
Therefore, the case in which both $u_1$ and $u_2$ are positive can never
happen.

\paragraph{Conclusion}
If $u_1$ and $u_2$ are negative, then the only generalized equilibrium is $(0,0)$.

Otherwise, $u_1 u_2 \leq 0$. For any pair of positive numbers $\alpha, \beta$
such that $\alpha u_1 + \beta u_2 = B$, there exists an equilbrium defined by
\eqref{eq:solution_equilibrium_x} and \eqref{eq:solution_equilibrium_y}. This
set of equilibria forms a half-line in $\cX \times \cY$ contained in the
hyperplane of points that saturate the constraint.

\paragraph{Special case: 1 dimension per player}
\[
u_1 = - \frac{c_x^2 + a c_x c_y}{1 + a^2}
\]
\[
u_2 = - \frac{c_y^2 - a c_x c_y}{1 + a^2} 
\]
If $a$ is very small, then we are in the empty case and there are no equilibria
that saturate the constraint.

\end{document}